\documentclass[11pt]{article}

\usepackage[final]{acl}

\usepackage{times}
\usepackage{latexsym}

\usepackage[T1]{fontenc}

\usepackage{microtype}

\newcommand{\cf}{\emph{cf. }}

\usepackage{inconsolata}

\usepackage{graphicx}
\usepackage{subcaption}
 \usepackage{balance}
\usepackage[table]{xcolor}
\usepackage[normalem]{ulem} 
\newcommand{\bestacc}[1]{\cellcolor{gray!12}\textbf{#1}}
\newcommand{\secondacc}[1]{\cellcolor{gray!8}\uline{#1}}
\newcommand{\forgetval}[1]{\textcolor{red!65!black}{#1}}
\newcommand{\posmark}[1]{\textcolor{green!50!black}{#1}}
\newcommand{\negmark}[1]{\textcolor{purple!60!black}{\textsf{x}}}

\usepackage{adjustbox}
\usepackage{booktabs}   
\usepackage{multirow}   
\usepackage{array}      
\usepackage{amsmath}      
\usepackage{amssymb}      
\usepackage{amsfonts}     

\title{PASs-MoE: Mitigating Misaligned Co-drift among Router and Experts via Pathway Activation Subspaces for Continual Learning}

\author{
\textbf{Zhiyan Hou}\textsuperscript{1,2}\!,
\textbf{Haiyun Guo}\textsuperscript{1,2}\thanks{Corresponding Authors}\!,
\textbf{Haokai Ma}\textsuperscript{3}\footnotemark[1]\!,
\textbf{Yandu Sun}\textsuperscript{4}\!,
\textbf{Yonghui Yang}\textsuperscript{3}\!,
\textbf{Jinqiao Wang}\textsuperscript{1,2,5}\\[0.4em]%
\textsuperscript{1}Institute of Automation, Chinese Academy of Sciences\quad
\textsuperscript{2}University of Chinese Academy of Sciences\\%
\textsuperscript{3}National University of Singapore\quad
\textsuperscript{4}Southeast University, Nanjing, China\\%
\textsuperscript{5}Wuhan AI Research, Wuhan, China\\[0.6em]%
{\small
\begin{minipage}{0.95\linewidth}\centering
houzhiyan23@mails.ucas.ac.cn\quad
haiyun.guo@nlpr.ia.ac.cn\quad
haokai.ma@nus.edu.sg\\
yh\_yang@nus.edu.sg\quad
jqwang@nlpr.ia.ac.cn
\end{minipage}}%
}

\begin{document}
\maketitle
\begin{abstract}
Continual instruction tuning (CIT) requires multimodal large language models (MLLMs) to adapt to a stream of tasks without forgetting prior capabilities. A common strategy is to isolate updates by routing inputs to different LoRA experts. However, existing LoRA-based Mixture-of-Experts (MoE) methods often jointly update the router and experts in an indiscriminate way, causing the router’s preferences to co-drift with experts’ adaptation pathways and gradually deviate from early-stage input–expert specialization. We term this as \emph{\textbf{Misaligned Co-drift}}, which blurs expert responsibilities and exacerbates forgetting.
To address this, we introduce the \emph{\textbf{pathway activation subspace (PASs)}}, a LoRA-induced subspace that reflects which low-rank pathway directions an input activates in each expert, providing a capability-aligned coordinate system for routing and preservation. Based on PASs, we propose a fixed-capacity PASs-based MoE–LoRA method with two components: PAS-guided Reweighting, which calibrates routing using each expert’s pathway activation signals, and PAS-aware Rank Stabilization, which selectively stabilizes rank directions important to previous tasks. Experiments on a CIT benchmark show that our approach consistently outperforms a range of conventional continual learning baselines and MoE–LoRA variants in both accuracy and anti-forgetting without adding parameters. Our code is publicly available at \url{https://github.com/yueluoshuangtian/PASs-MoE}.

\end{abstract}

\section{Introduction}
\label{sec:intro}

Multimodal large language models (MLLMs) have recently become a strong backbone for vision–language understanding and generation across diverse applications~\cite{achiam2023gpt, llava, liu2023visual, Qwen2.5VL,HyTuning,zhang2025cooper,hao2026clear}. However, task requirements and data distributions change over time in real-world deployments~\cite{ma2025attackseqbench}. Continual instruction tuning (CIT) studies how to sequentially adapt MLLMs on a stream of tasks with supervised instruction–response data while retaining previously acquired capabilities~\cite{Chung2022ScalingIL}.

\begin{figure}[t]
  \centering
  \includegraphics[width=0.49\textwidth]{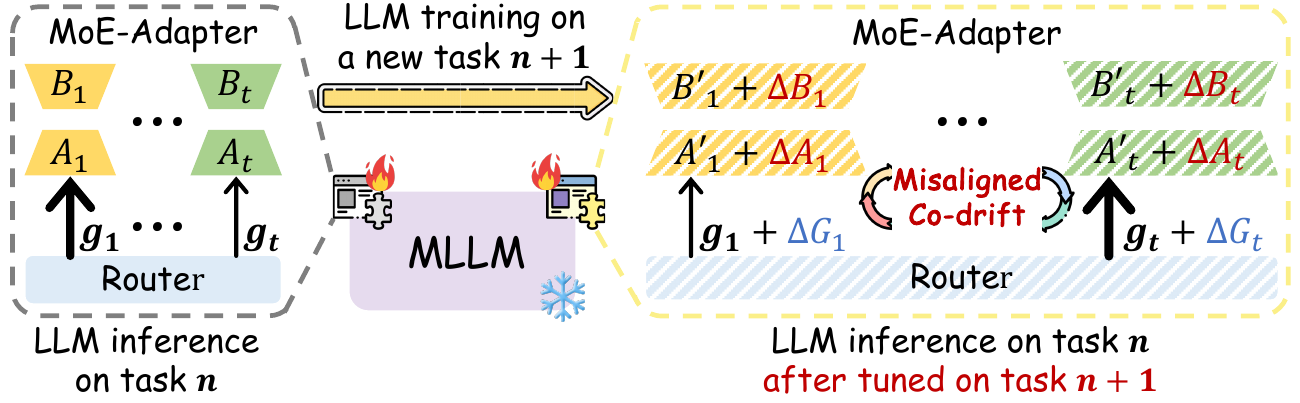}
  \vspace{-0.4cm}
  \caption{Illustration of the \emph{\textbf{Misaligned Co-drift}} in existing MoE-LoRA methods, where the thickness of arrow indicates the sampling probability of router. Here, the drift in router assignments and the internal drift within expert parameters for the same task jointly drive this issue, thereby exacerbating catastrophic forgetting.}
\vspace{-0.5cm}

  \label{fig:motivation}
\end{figure}

Serving as the most widely-used Parameter-efficient fine-tuning method for CIT~\cite{sung2022vl,zhang2021tip,hu2022lora}, the standard practice of sharing a single Low-rank adaptation (LoRA) module across tasks forces MLLMs to concentrate updates within the same low-rank subspace, increasing interference and exacerbating forgetting. To address this issue, recent studies combine the unique strengths of MoE architectures with LoRA. For instance, DDAS~\cite{yu2024boosting}, D-MoLE~\cite{ge2025dynamic} and RMoE~\cite{huai2025cl} attempt to expand MLLMs with additional experts and routing modules, which typically incur ever-growing parameters and storage, thereby increasing training and deployment complexity~\cite{fedus2022switch}. In contrast, MoELoRA~\cite{CoIN} routes each input to a fixed pool of LoRA experts, reducing forgetting during continual training without increasing model capacity. 
Nevertheless, its training procedure typically entails indiscriminate joint optimization of both the router and the expert parameters. As tasks progressively accumulate within the CIT sequence, both the router and expert parameters evolve to accommodate new tasks, yet their optimization directions are not fully aligned. This \emph{\textbf{Misaligned Co-drift}} may undermine the input-expert specialization established in earlier stages: \emph{Router updates can reroute the same input away from its previously associated expert, while expert updates alter that expert’s responses on samples from previous tasks.}

Drilling down into the underlying mechanism of \emph{\textbf{Misaligned Co-drift}}, we observe that existing MoE-LoRA variants fail to anchor routing and preservation in a capability-aligned coordinate system defined by each expert's low-rank adaptation pathway. Under this perspective, we revisit the formulation of MoE-LoRA~\cite{jacobs1991adaptive, shazeer2017outrageously}, especially the structural property of LoRA as $\Delta y(x)=B(Ax)$. Intuitively, the down-projection $A$ specifies the set of input directions an expert can respond to through its low-rank pathway, while the up-projection $B$ combines these coordinates into the additive correction within this low-rank coordinate system. Based on this observation, we creatively propose the notion of \emph{\textbf{P}athway \textbf{A}ctivation \textbf{S}ub\textbf{s}pace (\textbf{PASs})}  $\mathcal{S}=\mathrm{span}(A^\top)$. Unlike prior notions of ``activation subspace'' which are typically derived from intermediate activations, the proposed \emph{PASs} $\mathcal{S}$ is induced by LoRA’s down-projection $A$, where the low-rank activation $Ax$ provides a capability-tied coordinate system for aligning routing with parameter updates and selectively stabilizing task-critical rank directions, thus mitigating catastrophic forgetting in CIT.

Building on this notion, we propose a \textbf{PASs-based MoE-LoRA method} to mitigate catastrophic forgetting in CIT by aligning expert routing and preservation with each expert’s low-rank adaptation pathway. Our method consists of two components, \emph{PASs-guided Reweighting (PASs-RW)} and \emph{PASs-aware Rank Stabilization (PASs-RS)}, both grounded in the \emph{\textbf{PASs}}. Specifically, the proposed \emph{PASs-RW} modulates each expert’s contribution according to the input’s activation within the corresponding pathway activation subspace. Here, each expert’s routing weight is driven by its own low-rank pathway rather than by a separate router space, thereby alleviating \emph{misaligned co-drift}. Nevertheless, even with the improved routing alignment, the experts selected by the router still must adapt to new tasks, and their low-rank pathways may drift under the sequential updates, potentially leading to forgetting. To address this expert-side drift, we further design \emph{PASs-RS} to selectively constrain directions important to previous tasks by utilizing the same \emph{PASs} signal. This rank-level stabilization reduces forgetting while preserving plasticity.

Our contributions are threefold: (I) We introduce a \emph{pathway activation subspace} perspective for CL, modeling the low-dimensional update directions induced by LoRA as task-dependent pathway activation subspaces to analyze knowledge update and retention in MoE-LoRA. (\cf Section~\ref{sec:preliminary}) (II) We propose implicit self-routing and rank-level stability regularization for MoE-LoRA under a fixed parameter setting. (\cf Section~\ref{sec:methodology}) (III) Experiments demonstrate consistent improvements over a range of traditional CL algorithms and advanced MoE-related methods in both forgetting mitigation and downstream task performance (\cf Section~\ref{sec:experiments}).

\section{Related Works}
\paragraph{Continual Instruction Tuning.}
Continual Instruction Tuning (CIT) fine-tunes multimodal large language models (MLLMs)~\cite{zhang2025mm} over a task stream, training on task-specific instruction–response pairs while aiming to retain prior capabilities~\cite{he2023continual,CoIN,guo2025mllm,CPRec}. With limited or no access to past data, CIT faces the stability–plasticity trade-off and catastrophic forgetting. Existing mitigation strategies include importance constraints~\cite{Kirkpatrick_2017,li2017learning,mas}, relational knowledge distillation~\cite{zhang2024improving}, isolated adaptation~\cite{razdaibiedina2023progressive,l2p,cui2025cmoa,wang2023orthogonal}, and replay-based strategies~\cite{rolnick2019experience,replay,der,wang2024relational,he2024seekr,BalDRO} for approximating past distributions. CIT in MLLMs is further challenged by cross-modal alignment drift, which destabilizes vision-language mappings and exacerbates task interference~\cite{NS4RS,yang2026recall}.

\paragraph{Activation Subspace.}
Activation subspaces provide low-dimensional signals for interpreting and constraining model behavior~\cite{kim2018interpretability,yang2026revisiting}. CAV-style approaches represent concepts as activation directions~\cite{wenkmann2025variability,huang2024lg}, and subspace-based methods reduce interference by separating or constraining updates to task-relevant subspaces~\cite{farajtabar2020orthogonal,saha2021gradient,magistri2024elastic,roy2023subspace}, and low-dimensional projections of model representations have also proven effective for compressing reasoning traces into compact, task-aligned embeddings in multimodal models~\cite{hao2026trace, he2026plume}. These subspaces are typically statistics-driven and not tied to specific architectural units by design.. We instead define a LoRA-parameter-induced pathway activation subspace that binds subspace signals to low-rank channel functionality, enabling routing alignment and expert responsibility preservation.

\paragraph{Mixture of Experts.}
Sparse MoE has been explored for continual vision-language tuning, including MoE adapters and MoE-LoRA variants~\cite{CoIN,zhang2025enhancing,sun2025stronger} that dynamically compose LoRA experts, using rank-level micro-experts to retain pretraining knowledge~\cite{zhao2025each}.  However, most rely on explicitly defined expert pools and a shared routing space, which can accumulate misalignment between routing decisions and expert functionality. Our activation-subspace view couples routing with LoRA expert pathways to reduce routing drift and responsibility drift.

\begin{figure*}[t]
  \centering
  \includegraphics[width=0.98\linewidth]{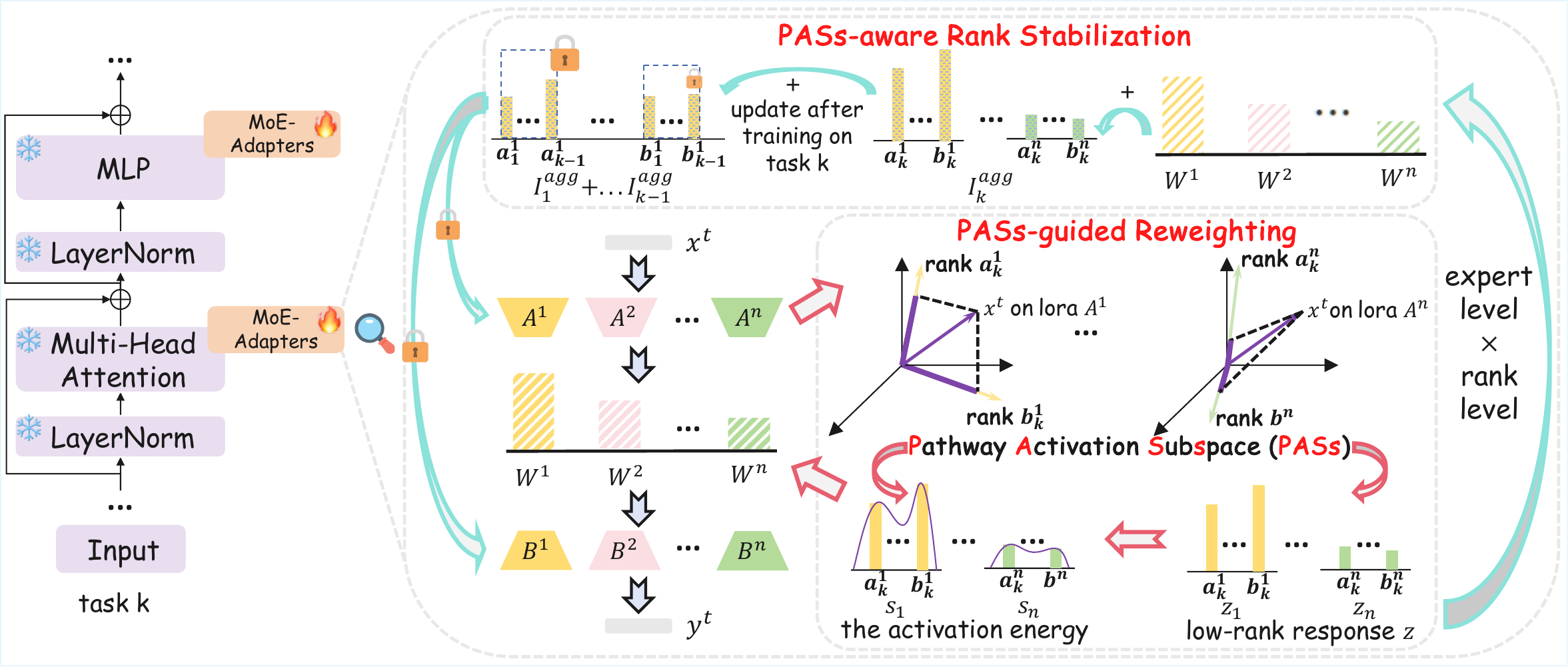}
  \vspace{-0.3cm}
\caption{\textbf{Overview of the proposed PASs-based MoE-LoRA method.} 
We consider continual instruction tuning with a fixed set of $E$ LoRA experts. 
(Bottom) Each input $x^t$ generates a low-rank response $z_e$ and activation energy $s_e$ via the Pathway Activation Subspace (PASs) of each expert. 
(Right) \textbf{PASs-guided Reweighting (PASs-RW)} leverages $s_e$ to compute mixture weights, eliminating the need for an independent router. 
(Top) \textbf{PASs-aware Rank Stabilization (PASs-RS)} tracks rank-level importance $I^{\mathrm{agg}}$ (represented by bar heights) and applies stabilization constraints (indicated by lock icons) to protect historically critical directions from drift.}
  \label{fig:overview}
  \vspace{-0.5cm}
\end{figure*}
\section{Preliminary}
\label{sec:preliminary}

\subsection{Problem Formulation}
\label{sec:setting}

We study continual instruction tuning (CIT) for Multimodal Large Language Models (MLLMs). Let $\{\mathcal{D}_t\}_{t=1}^{T}$ denote a stream of tasks, where each stage-$t$ dataset $\mathcal{D}_t=\{(v_i^{(t)}, x_i^{(t)}, y_i^{(t)})\}_{i=1}^{N_t}$ consists of multimodal instruction-response pairs. Here, $v_i^{(t)}$ denotes a visual input (e.g., an image), $x_i^{(t)}$ a textual instruction, and $y_i^{(t)}$ the target response. Starting from a pretrained multimodal backbone, training proceeds sequentially: at stage $t$, the model is updated using only the current dataset $\mathcal{D}_t$.

At each stage, the learning objective minimizes the empirical risk on the current task,
\begin{equation}
\label{eq:cit_obj}
\min_{\theta_t}\ \mathbb{E}_{(v,x,y)\sim \mathcal{D}_t}\big[\mathcal{L}(f(v,x;\theta_t), y)\big],
\end{equation}
where $f(\cdot;\theta_t)$ denotes the model at stage $t$. Following standard instruction tuning for MLLMs, we adopt the autoregressive negative log-likelihood loss. Under parameter-efficient fine-tuning, we let $\theta_t$ denote the trainable adapter parameters, while the pretrained backbone remains frozen.

\subsection{Decomposition of MoE-LoRA}
\label{sec:moelora}
Parameter-efficient fine-tuning (PEFT) adapts a pretrained model by introducing a small number of trainable parameters while keeping the backbone weights frozen~\cite{zhang2025parameter}. A widely used PEFT approach is Low-Rank Adaptation (LoRA), which augments a linear layer with a low-rank update. Given a weight matrix $W \in \mathbb{R}^{d_{\text{out}}\times d_{\text{in}}}$, LoRA defines
\begin{equation}
\label{eq:lora}
W' = W + \Delta W, \quad \Delta W = BA,
\end{equation}
where $A \in \mathbb{R}^{r\times d_{\text{in}}}$ and $B \in \mathbb{R}^{d_{\text{out}}\times r}$ are trainable matrices, and $r \ll \min(d_{\text{in}}, d_{\text{out}})$.

MoE-LoRA extends LoRA by instantiating multiple LoRA experts and combining their updates with a router. Consider $E$ experts $\{(A_e,B_e)\}_{e=1}^{E}$ attached to a linear layer, and let $h \in \mathbb{R}^{d_{\text{in}}}$ denote the hidden features input to this layer. The router $g(\cdot)$ maps $h$ to expert logits, from which mixture weights $\{\pi_e(h)\}_{e=1}^{E}$ are derived (e.g., dense normalization or sparse top-$k$ selection). The resulting input-dependent update to the weight matrix is
\begin{equation}
\label{eq:moelora}
\Delta W(h) = \sum_{e=1}^{E} \pi_e(h)\, B_e A_e h.
\end{equation}
In continual instruction tuning, PEFT optimizes the expert parameters $\{(A_e,B_e)\}_{e=1}^{E}$ together with the router parameters in $g$.

\subsection{Definition of Pathway Activation Subspace}
\label{sec:subspace}

We revisit the low-rank structure underlying MoE-LoRA. From the perspective of a single expert $e$, the LoRA branch adds an incremental output to the linear layer in the form
\begin{equation}
\Delta y_e(h)=B_e(A_e h).
\end{equation}
This factorization exposes an intermediate response vector $z_e(h)=A_e h$, which is a $r$-dimensional linear response induced by $A_e$ from the input $h$. The $j$-th entry $z_{e,j}(h)$ can be interpreted as the response coefficient along the $j$-th rank direction, capturing how strongly expert $e$ reacts to the input through that direction. The matrix $B_e$ then maps this $r$-dimensional response back to the output space, yielding the expert’s incremental contribution to the layer output.
Motivated by this structure, we define the PASs of expert $e$ as
\begin{equation}
\mathcal{S}_e=\mathrm{span}(A_e^\top).
\end{equation}
This subspace is induced by the row space of $A_e$ and characterizes the set of input-side directions that the expert uses to generate its low-rank response $z_e(h)$. In this sense, $\mathcal{S}_e$ provides a parameter-grounded reference space that directly links the expert’s input-side response pattern to its $A_e$. Building on this view, we develop PASs-RW and PASs-RS in Section~\ref{sec:methodology}.

\section{Methodology}
\label{sec:methodology}
\subsection{Overview}
\label{sec:method_overview}

We study continual instruction tuning (CIT) under a fixed-structure MoE-LoRA setting, where a single model learns tasks sequentially without expanding the expert pool.
Figure~\ref{fig:overview} illustrates the overall pipeline.
Given an input $x_t$, the backbone produces a hidden representation $h$.
Each LoRA expert $e$ defines a capability-tied pathway activation subspace (PASs) induced by its down-projection $A_e$, which yields a low-rank activation $z_e$.
We use this PASs signal as a shared anchor to jointly design routing and preservation, targeting the \emph{Misaligned Co-drift} discussed in the introduction.

Based on this view, we propose two complementary and synergistic components. First, \textbf{PASs-RW} computes mixture weights from the expert-specific low-rank activation, so that expert assignment is driven by the expert pathway itself rather than by an independent router space. Second, \textbf{PASs-RS} dynamically accumulates PASs-based activation statistics during sequential training and selectively stabilizes historically important rank directions.

\begin{table*}[t]
\centering
\begin{adjustbox}{max width=\textwidth}
{\fontsize{9pt}{10pt}\selectfont
\setlength{\tabcolsep}{1mm}
\begin{tabular}{l
                cc cc cc cc cc cc cc
                cc}
\toprule
\multirow{2}{*}{\textbf{Method}} &
\multicolumn{2}{c}{\textbf{Math QA}} &
\multicolumn{2}{c}{\textbf{Arts VQA}} &
\multicolumn{2}{c}{\textbf{Math VQA}} &
\multicolumn{2}{c}{\textbf{Econ.\ QA}} &
\multicolumn{2}{c}{\textbf{Med.\ VQA}} &
\multicolumn{2}{c}{\textbf{OCR VQA}} &
\multicolumn{2}{c}{\textbf{Sci.\ VQA}} &
\multirow{2}{*}{\textbf{AP}} & \multirow{2}{*}{\textbf{BWT}} \\
\cmidrule(lr){2-3}\cmidrule(lr){4-5}\cmidrule(lr){6-7}\cmidrule(lr){8-9}
\cmidrule(lr){10-11}\cmidrule(lr){12-13}\cmidrule(lr){14-15}
& Acc & Forget & Acc & Forget & Acc & Forget & Acc & Forget
& Acc & Forget & Acc & Forget & Acc & Forget &  &  \\
\midrule

\textbf{SeqFT} &
\secondacc{37.89} & \forgetval{$-19.75$} &
5.52  & \forgetval{$-32.35$} &
39.91 & \forgetval{$-11.97$} &
\secondacc{64.42} & \forgetval{$-5.74$} &
15.70 & \forgetval{$-20.40$} &
7.19  & \forgetval{$-16.67$} &
\bestacc{84.07} & -- &
36.39 & \forgetval{$-15.27$} \\

\textbf{SeqLoRA} &
0.00  & \forgetval{$-54.43$} &
6.28  & \forgetval{$-24.81$} &
27.75 & \forgetval{$-22.18$} &
38.44 & \forgetval{$-35.15$} &
24.70 & \forgetval{$-12.54$} &
16.96 & \forgetval{$-5.53$} &
83.52 & -- &
28.24 & \forgetval{$-22.09$} \\

\textbf{Replay} &
40.64 & \forgetval{$-12.56$} &
25.44 & \forgetval{$-7.97$} &
34.37 & \forgetval{$-14.09$} &
60.15 & \forgetval{$-11.83$} &
\bestacc{31.79} & \forgetval{$-2.34$} &
\bestacc{23.68} & \forgetval{$-3.21$} &
78.25 & --  &
42.05 & \forgetval{$-8.29$} \\

\textbf{L2P} &
17.54 & \forgetval{$-26.95$} &
16.06  & \forgetval{$-19.58$} &
29.41 & \forgetval{$-17.32$} &
42.57 & \forgetval{$-29.48$} &
21.27 & \forgetval{$-6.48$} &
16.45 & \forgetval{$-4.97$} &
78.25 & -- &
31.65 & \forgetval{$-14.97$} \\

\textbf{EWC} &
16.78 & \forgetval{$-36.18$} &
6.83  & \forgetval{$-26.42$} &
25.68 & \forgetval{$-26.66$} &
37.40 & \forgetval{$-36.19$} &
23.63 & \forgetval{$-11.25$} &
17.49 & \forgetval{$-4.66$} &
83.52 & -- &
30.19 & \forgetval{$-20.20$} \\

\textbf{LwF} &
18.98 & \forgetval{$-34.71$} &
6.45  & \forgetval{$-27.01$} &
26.80 & \forgetval{$-25.31$} &
36.51 & \forgetval{$-34.66$} &
23.61 & \forgetval{$-9.94$} &
17.38 & \forgetval{$-8.10$} &
81.68 & -- &
30.20 & \forgetval{$-19.96$} \\

\textbf{MAS} &
19.48 & \forgetval{$-34.95$} &
7.87  & \forgetval{$-25.65$} &
28.48 & \forgetval{$-23.63$} &
38.48 & \forgetval{$-35.11$} &
21.58 & \forgetval{$-13.37$} &
17.78 & \forgetval{$-6.80$} &
78.74 & -- &
30.34 & \forgetval{$-19.36$} \\

\textbf{O-LoRA} &
24.58 & \forgetval{$-29.85$} &
9.87  & \forgetval{$-22.61$} &
30.79 & \forgetval{$-20.98$} &
38.48 & \forgetval{$-34.01$} &
22.82 & \forgetval{$-12.85$} &
18.64 & \forgetval{$-5.92$} &
82.44 & -- &
32.52 & \forgetval{$-19.46$} \\

\textbf{HiDe-LLaVA} &
26.85 & \forgetval{$-25.12$} &
11.84 & \forgetval{$-22.32$} &
31.58 & \forgetval{$-16.77$} &
49.29 & \forgetval{$-9.08$} &
8.22  & \forgetval{$-20.75$} &
14.67 & \forgetval{$-3.07$} &
29.97 & -- &
24.63 & \forgetval{$-16.16$} \\

\midrule 

\rowcolor{gray!6}
\textbf{DDAS} &
\secondacc{44.83} & \forgetval{$-7.88$} &
20.87 & \forgetval{$-5.36$} &
37.40 & \forgetval{$-1.25$} &
58.77 & \forgetval{$-13.51$} &
23.74 & \forgetval{$-0.34$} &
2.35  & \forgetval{$-0.07$} &
74.46 & -- &
37.49 & \forgetval{$-4.06$} \\

\rowcolor{gray!6}
\textbf{MoELoRA (Top-k)} &
35.68 & \forgetval{$-15.06$} &
6.78  & \forgetval{$-24.18$} &
33.58 & \forgetval{$-17.69$} &
39.80 & \forgetval{$-27.18$} &
22.67 & \forgetval{$-10.89$} &
14.57 & \forgetval{$-3.41$} &
73.56 & -- &
32.38 & \forgetval{$-15.34$} \\

\rowcolor{gray!6}
\textbf{MoELoRA (Softmax)} &
42.98 & \forgetval{$-11.45$} &
\secondacc{35.89} & $+2.65$ &
\secondacc{40.84} & \forgetval{$-11.61$} &
56.10 & \forgetval{$-15.47$} &
29.24 & \forgetval{$-6.04$} &
18.89 & \forgetval{$-4.57$} &
79.57 & -- &
\secondacc{43.36} & \forgetval{$-6.64$} \\

\rowcolor{gray!10}
\textbf{Ours} &
\bestacc{49.52} & \forgetval{$-5.90$} &
\bestacc{43.22} & $+10.43$ &
\bestacc{44.70} & \forgetval{$-6.50$} &
\bestacc{66.13} & \forgetval{$-6.05$} &
\secondacc{29.99} & \forgetval{$-5.38$} &
\secondacc{21.95} & \forgetval{$-1.63$} &
\secondacc{83.73} & -- &
\bestacc{48.46} & \forgetval{$-2.15$} \\

\bottomrule
\end{tabular}}
\end{adjustbox}
\caption{Comparison with traditional methods and MoE-LoRA-based methods on \textsc{MLLM-CTBench}. Best and second-best results for Acc and AP are marked in \textbf{bold} and \underline{underline}.}

\label{tab:main_continual_results}
\end{table*}

\subsection{PASs-guided Reweighting}
\label{sec:pasrw}

PASs-RW mitigates \emph{Misaligned Co-drift} by tying routing to each expert's functional response. Prior MoE-LoRA variants often learn routing in a shared space that is decoupled from the experts' low-rank pathways, so routing policies and expert parameters can drift inconsistently as sequential tasks accumulate. We instead derive routing weights from the low-rank activation $z_e=A_e h$, which measures how strongly expert $e$ responds specifically to the input along its LoRA pathway.

We define the activation energy of expert $e$ as
\begin{equation}
s_e(h)=\frac{1}{r}\lVert A_e h\rVert_2^2,
\end{equation}
and compute mixture coefficients by softmax normalization:
\begin{equation}
\pi_e(h)=\frac{\exp(s_e(h))}{\sum_{e'=1}^{E}\exp(s_{e'}(h))}.
\end{equation}
Grounding $\pi_e(h)$ in $z_e$ anchors routing in a capability-tied signal, coupling routing behavior with expert evolution under sequential training.

\subsection{PASs-aware Rank Stabilization}
\label{sec:pasrs}

While PASs-RW couples routing to the expert pathway, the selected experts still need to adapt to new tasks, and their low-rank parameters may drift under sequential updates, which can induce forgetting. The PAS formulation makes this drift analyzable at the rank level. Specifically, we define the low-rank response as $z_e = A_e h$, where $z_e \in \mathbb{R}^{r}$, which admits a coordinate-wise interpretation: each coordinate corresponds to the activation of one rank direction of $A_e$. Let $a_{e,k}\in\mathbb{R}^{d}$ denote the $k$-th row of $A_e$, then the $k$-th component of $z_e$ is
\begin{equation}
z_{e,k} \;=\; a_{e,k}^{\top} h .
\end{equation}
This view enables rank-level importance estimation within each expert. We define the importance of rank direction $(e,k)$ on task $t$ as
\begin{equation}
\begin{aligned}
I_{e,k}(t)
&\triangleq \mathbb{E}_{h\sim \mathcal{D}_t}
\Big[\pi_e(h)\,\big(z_{e,k}\big)^2\Big] \\
&= \mathbb{E}_{h\sim \mathcal{D}_t}
\Big[\pi_e(h)\,(a_{e,k}^\top h)^2\Big].
\end{aligned}
\end{equation}
We maintain the aggregated importance over past tasks to track historical significance as
\begin{equation}
I_{e,k}^{\mathrm{agg}}(t-1)\triangleq \sum_{t'=1}^{t-1} I_{e,k}(t').
\end{equation}
Let $b_{e,k}\in\mathbb{R}^{d_{\text{out}}}$ denote the $k$-th column of $B_e$. Since drift in either the input-side factor $a_{e,k}$ or the output-side factor $b_{e,k}$ can alter expert behavior along rank direction $k$, we regularize both factors using the same PASs-derived importance weights.

When learning task $t$, we impose a weighted stabilization constraint on historically important rank directions to mitigate catastrophic forgetting:

\begin{equation}
\begin{aligned}
\mathcal{L}_{\mathrm{stab}}^{A}
&=\sum_{e=1}^{E}\sum_{k=1}^{r} w_{e,k}\,
\bigl\lVert a_{e,k}^{(t)}-a_{e,k}^{(t-1)}\bigr\rVert_2^2, \\
\mathcal{L}_{\mathrm{stab}}^{B}
&=\sum_{e=1}^{E}\sum_{k=1}^{r} w_{e,k}\,
\bigl\lVert b_{e,k}^{(t)}-b_{e,k}^{(t-1)}\bigr\rVert_2^2 .
\end{aligned}
\label{eq:stab}
\end{equation}
where $w_{e,k}$ is a normalized and clipped version of the aggregated importance
$I_{e,k}^{\mathrm{agg}}(t\!-\!1)$ to ensure scale invariance and numerical stability.
The overall objective is $\mathcal{L}=\mathcal{L}_{\mathrm{task}}+\lambda_A \mathcal{L}_{\mathrm{stab}}^{A}+\lambda_B \mathcal{L}_{\mathrm{stab}}^{B}$. By concentrating stabilization on rank directions that are more important to past tasks, PASs-RS reduces expert-side drift that harms prior capabilities. Here, $A$ serves as the structural anchor of the pathway activation subspace, while moderate regularization on the paired columns of $B$ provides complementary robustness under multi-task interference. Moreover, since PASs-RW derives routing weights from the same pathway response, stabilizing task-critical directions in $A$ also helps maintain more consistent routing for earlier-task inputs during later training.

\begin{table*}[t]
\centering
\begin{adjustbox}{max width=\textwidth}
{\fontsize{9pt}{10pt}\selectfont
\setlength{\tabcolsep}{1mm}
\renewcommand{\arraystretch}{1.0}
\begin{tabular}{c c cc
                cc cc cc cc cc cc cc
                cc}
\hline
\multirow{2}{*}{\textbf{Method}} &
\multirow{2}{*}{\textbf{MoE}} &
\multirow{2}{*}{\textbf{PASs-RW}} &
\multirow{2}{*}{\textbf{PASs-RS}} &
\multicolumn{2}{c}{\textbf{Math QA}} &
\multicolumn{2}{c}{\textbf{Arts VQA}} &
\multicolumn{2}{c}{\textbf{Math VQA}} &
\multicolumn{2}{c}{\textbf{Econ.\ QA}} &
\multicolumn{2}{c}{\textbf{Med.\ VQA}} &
\multicolumn{2}{c}{\textbf{OCR VQA}} &
\multicolumn{2}{c}{\textbf{Sci.\ VQA}} &
\multirow{2}{*}{\textbf{AP}} & \multirow{2}{*}{\textbf{BWT}} \\
\cmidrule(lr){5-6}\cmidrule(lr){7-8}\cmidrule(lr){9-10}\cmidrule(lr){11-12}\cmidrule(lr){13-14}\cmidrule(lr){15-16}\cmidrule(lr){17-18}
& & & &
Acc & Forget & Acc & Forget & Acc & Forget & Acc & Forget
& Acc & Forget & Acc & Forget & Acc & Forget &  &  \\ \midrule


\rowcolor{gray!6}
\textbf{MoELoRA (top-$k$)} & Top-$k$ & \negmark{x} & \negmark{x} &
35.68 & \forgetval{$-15.06$} &
6.78  & \forgetval{$-24.18$} &
33.58 & \forgetval{$-17.69$} &
39.80 & \forgetval{$-27.18$} &
22.67 & \forgetval{$-10.89$} &
14.57 & \forgetval{$-3.41$} &
73.56 & -- &
32.38 & \forgetval{$-15.34$} \\[4pt]

\rowcolor{gray!6}
\textbf{MoELoRA (softmax)} & All & \negmark{x} & \negmark{x} &
42.98 & \forgetval{$-11.45$} &
35.89 & $+2.65$ &
40.84 & \forgetval{$-11.61$} &
56.10 & \forgetval{$-15.47$} &
29.24 & \forgetval{$-6.04$} &
18.89 & \forgetval{$-4.57$} &
79.57 & -- &
43.36 & \forgetval{$-6.64$} \\[4pt]

\rowcolor{gray!6}
\textbf{+ PASs-RW} & All & \posmark{\checkmark} & \negmark{x} &
\secondacc{49.99} & \forgetval{$-5.18$} &
\secondacc{39.11} & $+6.26$ &
\secondacc{42.70} & \forgetval{$-8.27$} &
\secondacc{60.90} & \forgetval{$-9.87$} &
\secondacc{27.79} & \forgetval{$-8.74$} &
\secondacc{20.27} & \forgetval{$-3.86$} &
\secondacc{83.41} & -- &
\secondacc{46.31} & \forgetval{$-4.24$} \\[2pt]

\rowcolor{gray!10}
\textbf{Ours (Full)} & All & \posmark{\checkmark} & \posmark{\checkmark} &
\bestacc{49.52} & \forgetval{$-5.90$} &
\bestacc{43.22} & $+10.43$ &
\bestacc{44.70} & \forgetval{$-6.50$} &
\bestacc{66.13} & \forgetval{$-6.05$} &
\bestacc{29.99} & \forgetval{$-5.38$} &
\bestacc{21.95} & \forgetval{$-1.63$} &
\bestacc{83.73} & -- &
\bestacc{48.46} & \forgetval{$-2.15$} \\[2pt]

\hline
\end{tabular}}
\end{adjustbox}
\caption{
Structural ablation of our method on continual instruction learning. We compare two MoE-LoRA baselines, using either softmax weighting over all experts (All) or Top-$k$ expert selection, and then progressively add PASs-RW and PASs-RS, where the combination corresponds to our method. Best and second-best results for Acc and AP are marked in \textbf{bold} and \uline{underline}.}
\label{tab:ablation_moe}
\end{table*}

\section{Experiments}
\label{sec:experiments}
\subsection{Experimental Setting}
\label{sec:exp_setup}
\paragraph{Datasets}
We evaluate our method on MLLM-CTBench, a continual instruction tuning benchmark for multimodal large language models. Following the official protocol, we perform sequential training on the default task sequence without revisiting past data. MLLM-CTBench involves heterogeneous tasks with substantial distribution shifts, posing a challenging testbed for continual adaptation. Detailed dataset descriptions and the exact evaluation protocol are provided in Appendix~\ref{app:setup}.

\paragraph{Baselines}
We compare against representative continual learning methods for MLLMs, including LwF~\cite{lwf}, EWC~\cite{ewc}, L2P~\cite{l2p}, O-LoRA~\cite{wang2023orthogonal}, and HiDe-LLaVA~\cite{guo2025hide}, and also include MoE-LoRA variants (MoELoRA and DDAS). Full training recipes and hyperparameters are provided in Appendix~\ref{app:setup}.

\paragraph{Implementation Details} We adopt LLaVA-v1.5-7B~\cite{llava} as the base model and use CLIP-L/14-336~\cite{radford2021learning} to extract visual and textual features. Following LLaVA’s LoRA protocol, we add LoRA to all language-model linear layers with rank $r{=}128$. Unless otherwise specified, we use $E=6$ experts and set the stabilization coefficients to $\lambda_A=5\times10^{-4}$ for LoRA $A$ and $\lambda_B=5\times10^{-5}$ for LoRA $B$. Under the continual learning setting, we perform sequential training strictly following the task order of each benchmark, with 3 epochs per task on \textsc{MLLM-CTBench}, using a warmup ratio of 0.03. All methods are trained with a batch size of 12 on NVIDIA H20 GPUs, and all experiments are conducted with a fixed random seed of 42.

\paragraph{Metrics}

We report the final performance on each task $t$ after training on the full stream,
$\mathrm{Acc}_t^{\text{final}}$, and the corresponding signed task-wise forgetting
$\mathrm{Forget}_t=\mathrm{Acc}_t^{\text{final}}-\mathrm{Acc}_t^{\text{after }t}$,
where $\mathrm{Acc}_t^{\text{after }t}$ is the accuracy measured immediately after learning task $t$.
We summarize overall continual learning performance with
$\mathrm{AP}=\frac{1}{T}\sum_{t=1}^{T}\mathrm{Acc}_t^{\text{final}}$ and
$\mathrm{BWT}=\frac{1}{T}\sum_{t=1}^{T}\mathrm{Forget}_t$.
Under this convention, higher $\mathrm{Forget}_t$/$\mathrm{BWT}$ indicates less forgetting; when values are negative, those closer to zero are better.

\subsection{Main Results}
\label{sec:main_results}

The main results on \textsc{MLLM-CTBench} are reported in Table~\ref{tab:main_continual_results}, while results on an alternative task order are provided in Appendix~\ref{app:order}. Our method achieves the best overall performance, improving the final Average Performance (AP) and reducing forgetting measured by Backward Transfer (BWT) compared with conventional continual learning baselines. It also surpasses MoE-LoRA variants, including the two representative design lines summarized in the Introduction. Notably, on \textsc{MLLM-CTBench}, our final AP exceeds that of the second-best method by \textbf{5.1\%}. We observe consistent gains under the alternative task order, where our AP improves by \textbf{9.46\%} over the second-best method (Appendix~\ref{app:order}).

\subsection{Ablation Study}
\label{sec:ablation}
Table~\ref{tab:ablation_moe} reports the structural ablation results under a fixed MoE-LoRA parameter budget. 
Since the Top-$k$ routing baseline performs substantially worse in the continual instruction tuning setting, we treat it only as a reference point. 
Accordingly, we introduce our designs on top of the softmax-based MoE-LoRA baseline and progressively ablate the two core components of our framework.
\subsubsection{Structural Ablation}
Table~\ref{tab:ablation_moe} presents structural ablations under a fixed MoE-LoRA parameter budget. Since the Top-$k$ routing baseline performs notably worse in continual instruction tuning, we include it only as a reference point. All ablations are built upon a softmax-based MoE-LoRA baseline.

We ablate the two key components of our framework to quantify their individual contributions. Enabling \emph{PASs-RW} alone consistently improves the final performance over the MoE-LoRA baseline, suggesting that activation-subspace signals provide an effective basis for expert reweighting under sequential training. This observation aligns with our motivation in the Introduction that routing should be grounded in expert-specific low-rank pathways, rather than learned in a separate shared routing space. We defer a closer examination of routing dynamics to the dedicated analysis section~\ref{sec:analysis}.

\begin{figure}[t]
    \centering
    \includegraphics[width=\linewidth]{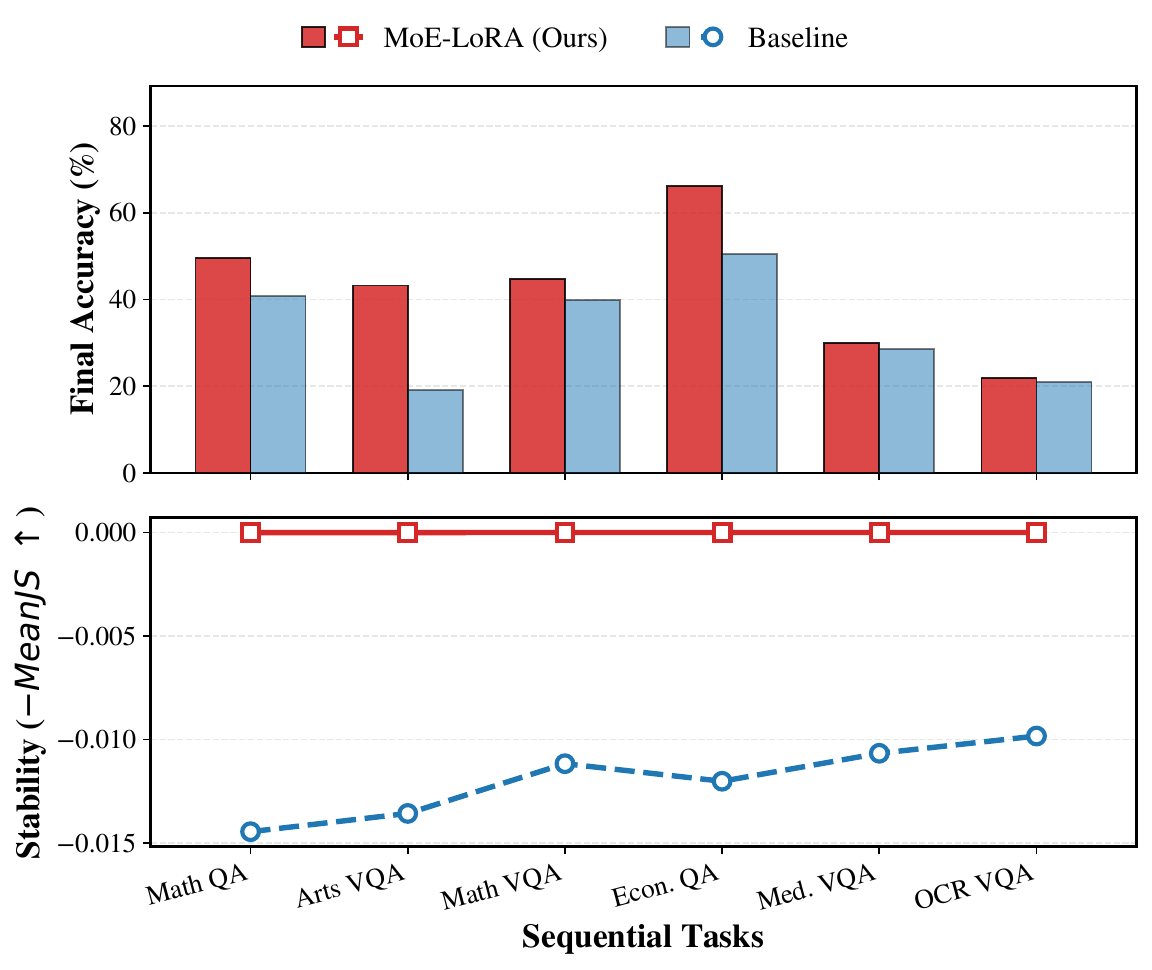}
    \caption{
    \textbf{Comparison of final performance and router stability.} 
    Top and bottom panels display the final accuracy and the corresponding router stability across the task stream, respectively. Stability is quantified as the negative Mean Jensen--Shannon (JS) divergence.
    }
    \label{fig:acc-vs-stability}
\end{figure}
Adding \emph{PASs-aware Rank Stabilization (PASs-RS)} on top of reweighting yields additional gains, indicating that selectively constraining task-relevant rank directions can further improve retention while preserving adaptation to new tasks. We further compare our stabilization with a random regularization scheme; the corresponding results are reported in Appendix~\ref{app:extra}. Overall, the comparison consistently supports the essential importance of using these activation-subspace signals to judiciously decide which rank directions to stabilize.

\subsubsection{Hyperparameter Ablation}
\begin{figure}[t]
    \centering
    \includegraphics[width=\columnwidth]{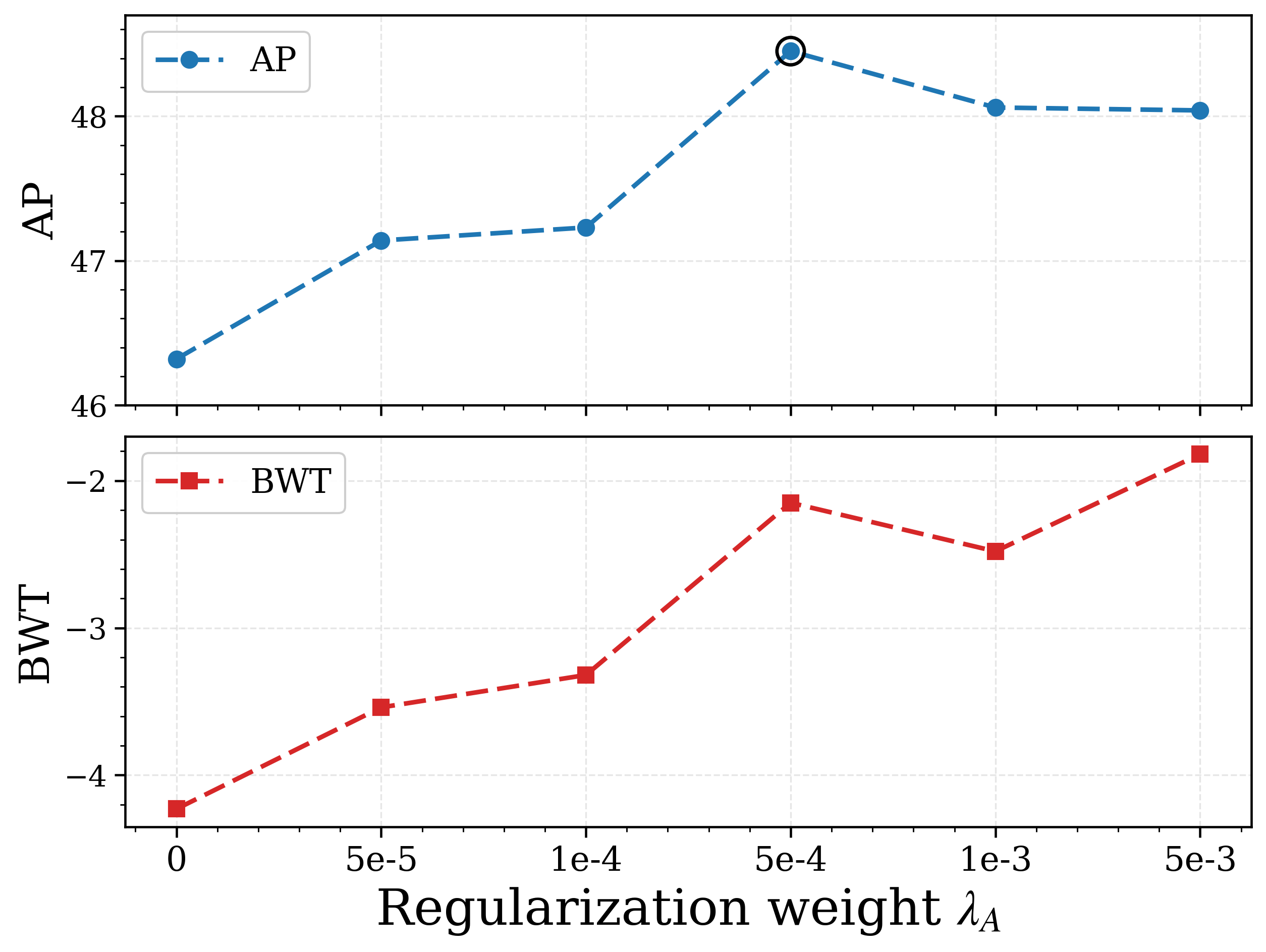}
    \caption{
        Ablation results on the regularization strength $\lambda_A$ for PASs-MoE in
        MLLM-CTBench. The upper panel reports AP, while the lower panel shows BWT.
    }
    \label{fig:ablation_regularization}
\end{figure}

\begin{figure}[t]
    \centering
    \includegraphics[width=\columnwidth]{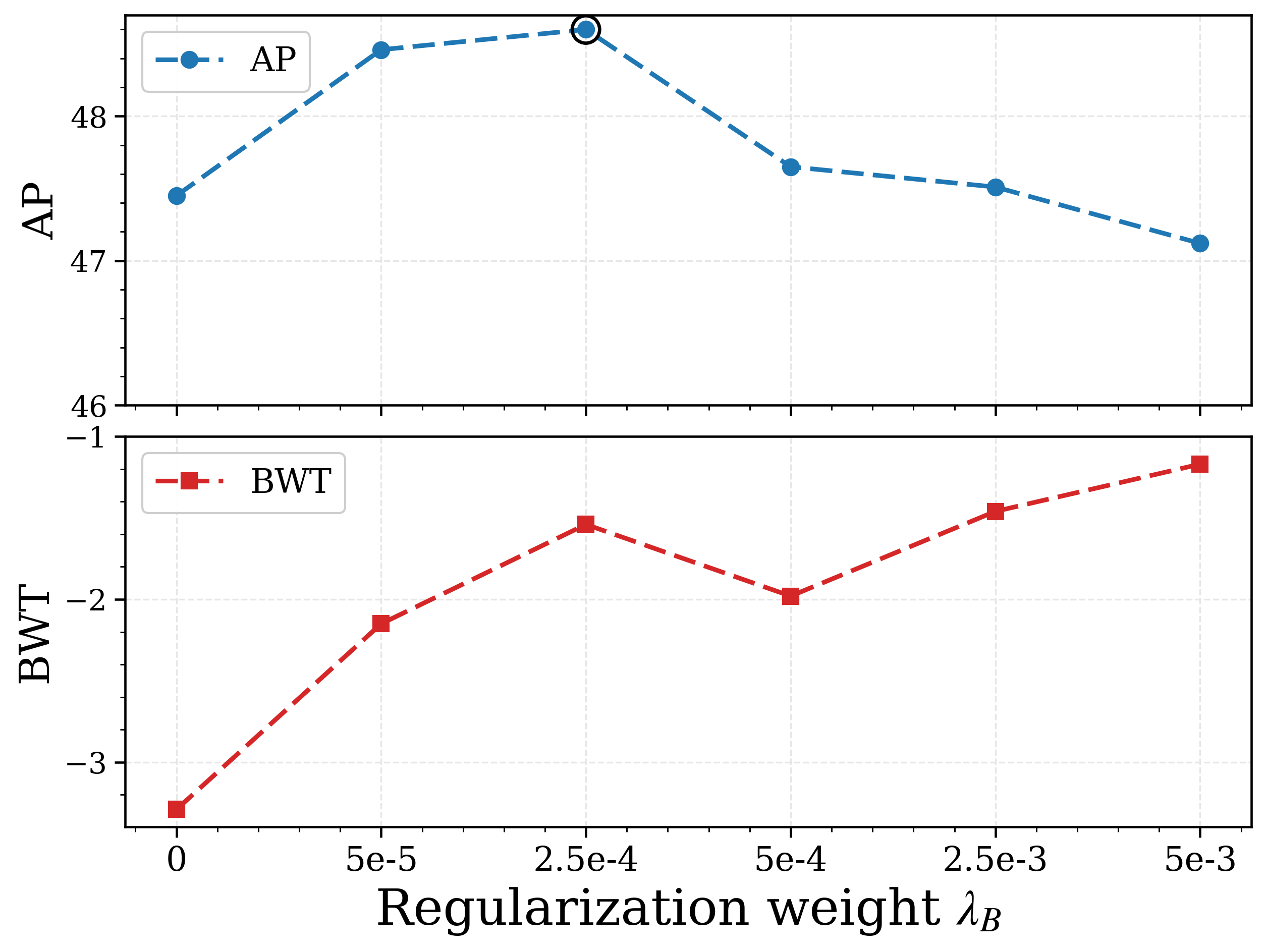}
    \caption{
        Ablation results on the regularization strength $\lambda_B$ for PASs-MoE in
        MLLM-CTBench. The upper panel reports AP, while the lower panel shows BWT.
    }
    \label{fig:ablation_regularization_B}
\end{figure}

We further study sensitivity to three key hyperparameters: the number of experts $E$ (Table~\ref{tab:ablation_num_experts}), the stabilization strength $\lambda_A$ on LoRA $A$ (Fig.~\ref{fig:ablation_regularization}), and the stabilization strength $\lambda_B$ on LoRA $B$ (Fig.~\ref{fig:ablation_regularization_B}). 
For the expert count, the performance is non-monotonic: our method achieves the best final AP at a moderate $E$ (peaking at $E{=}6$), while larger $E$ brings marginal or slightly reduced gains (e.g., $E{=}8$). This trend suggests that there exists a favorable capacity regime under a fixed training budget, whereas further increasing $E$ may yield diminishing returns (Table~\ref{tab:ablation_num_experts}).

For the stabilization strength $\lambda_A$, we observe a clear stability-plasticity trade-off (Fig.~\ref{fig:ablation_regularization}).
As $\lambda_A$ increases, forgetting is gradually alleviated, reflected by the monotonic increase of BWT.
Meanwhile, the final performance (AP) exhibits a non-monotonic trend: it improves under mild regularization but starts to drop when $\lambda_A$ becomes larger.
This is expected because stronger regularization increasingly constrains parameter updates, reducing the model's ability to adapt to new tasks; as a result, forgetting is reduced at the cost of diminished task learning, leading to lower AP at high $\lambda_A$.

For the output-side stabilization strength $\lambda_B$, we fix $\lambda_A=5\times10^{-4}$ and vary $\lambda_B=\alpha\lambda_A$ (Fig.~\ref{fig:ablation_regularization_B}). Regularizing $B$ provides complementary benefits for reducing forgetting, but is more sensitive to the choice of $\alpha$ than stabilizing $A$. The best trade-off is obtained at $\alpha=0.5$ ($\lambda_B=2.5\times10^{-4}$), where AP reaches its peak while BWT is substantially improved. Larger values (e.g., $\alpha\geq 1.0$) tend to over-constrain plasticity and slightly degrade final AP, even though BWT can continue to improve. Overall, these results suggest that $A$ serves as the more stable structural anchor, while moderate $B$ regularization further improves robustness under multi-task interference.

\begin{figure*}[t]
  \centering
  \begin{subfigure}[t]{0.32\linewidth}
    \centering
    \includegraphics[width=\linewidth]{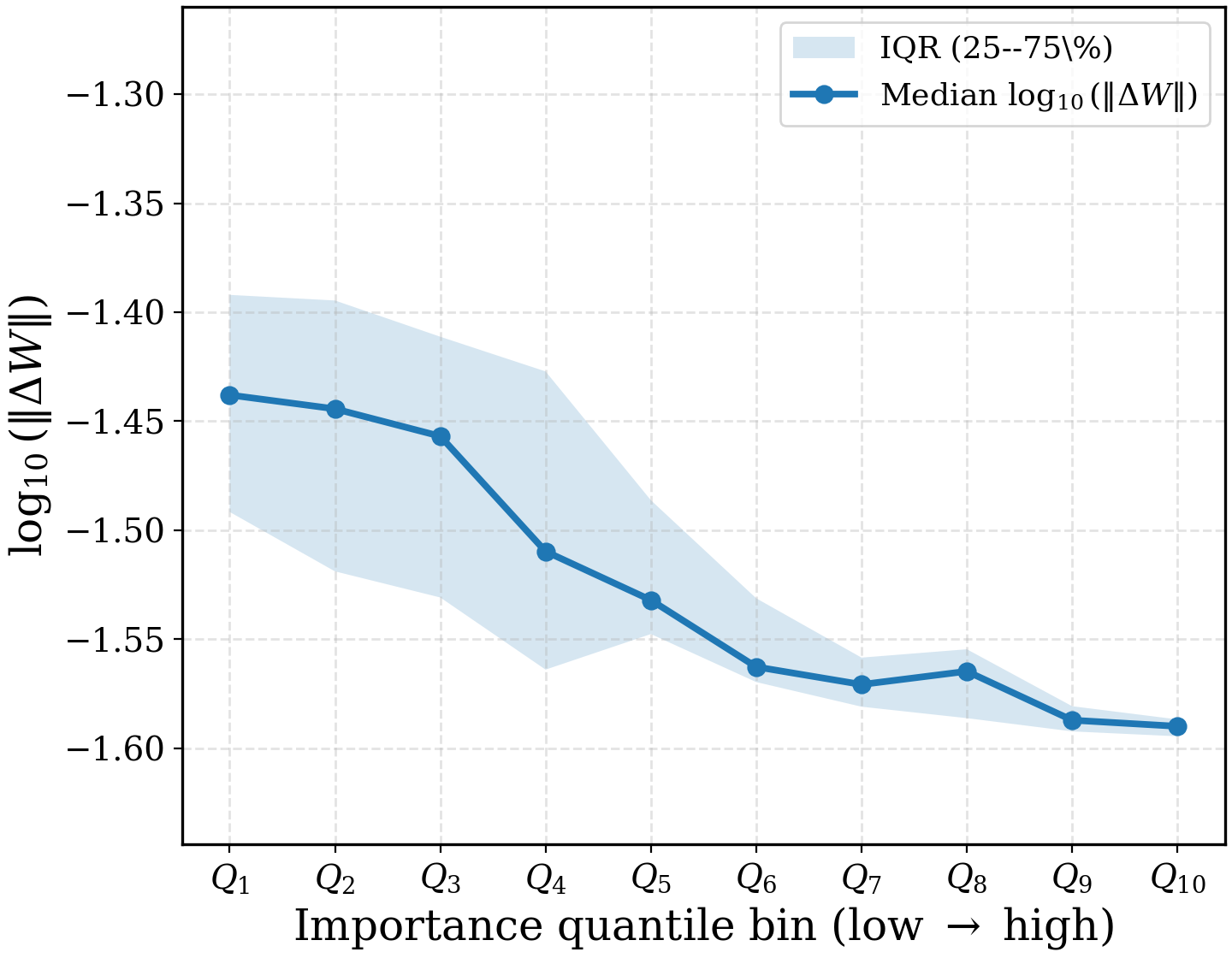}
    \caption{Layer 1 (low-level)}
    \label{fig:upd_imp:l1}
  \end{subfigure}
  \hfill
  \begin{subfigure}[t]{0.32\linewidth}
    \centering
    \includegraphics[width=\linewidth]{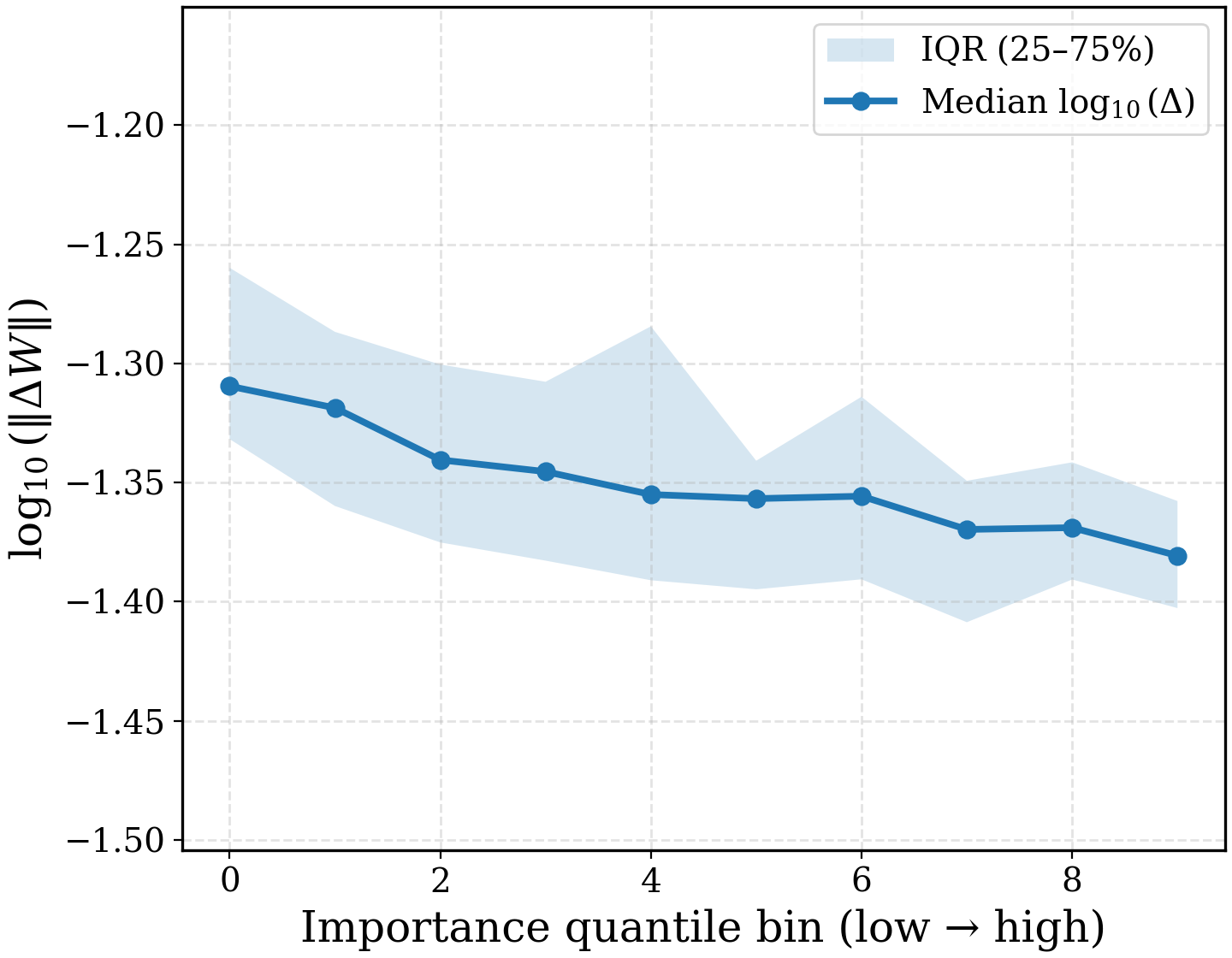}
    \caption{Layer 15 (mid-level)}
    \label{fig:upd_imp:l15}
  \end{subfigure}
  \hfill
  \begin{subfigure}[t]{0.32\linewidth}
    \centering
    \includegraphics[width=\linewidth]{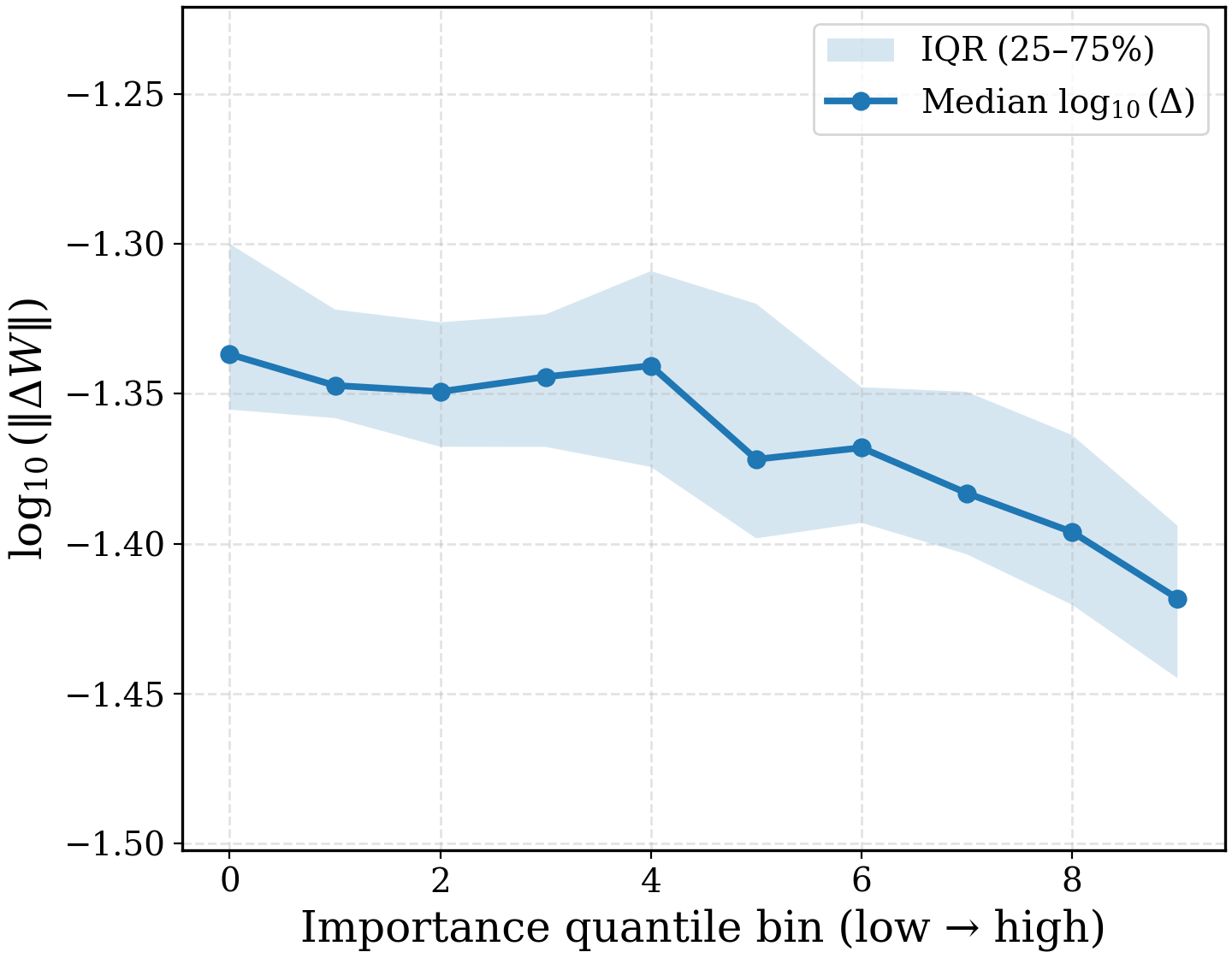}
    \caption{Layer 29 (top-level)}
    \label{fig:upd_imp:l29}
  \end{subfigure}
  \caption{\textbf{Update magnitude versus old-task importance across representative layers.}
  Each plot bins coefficients by the previous task's aggregated importance $I_{\mathrm{agg}}$ (low $\rightarrow$ high) and reports the median (with interquartile range) of the log update norm $\log \|\Delta W\|$ within each bin.}
  \label{fig:upd_imp:repr}
\end{figure*}

\begin{figure}[t]
  \centering
  \includegraphics[width=0.48\textwidth]{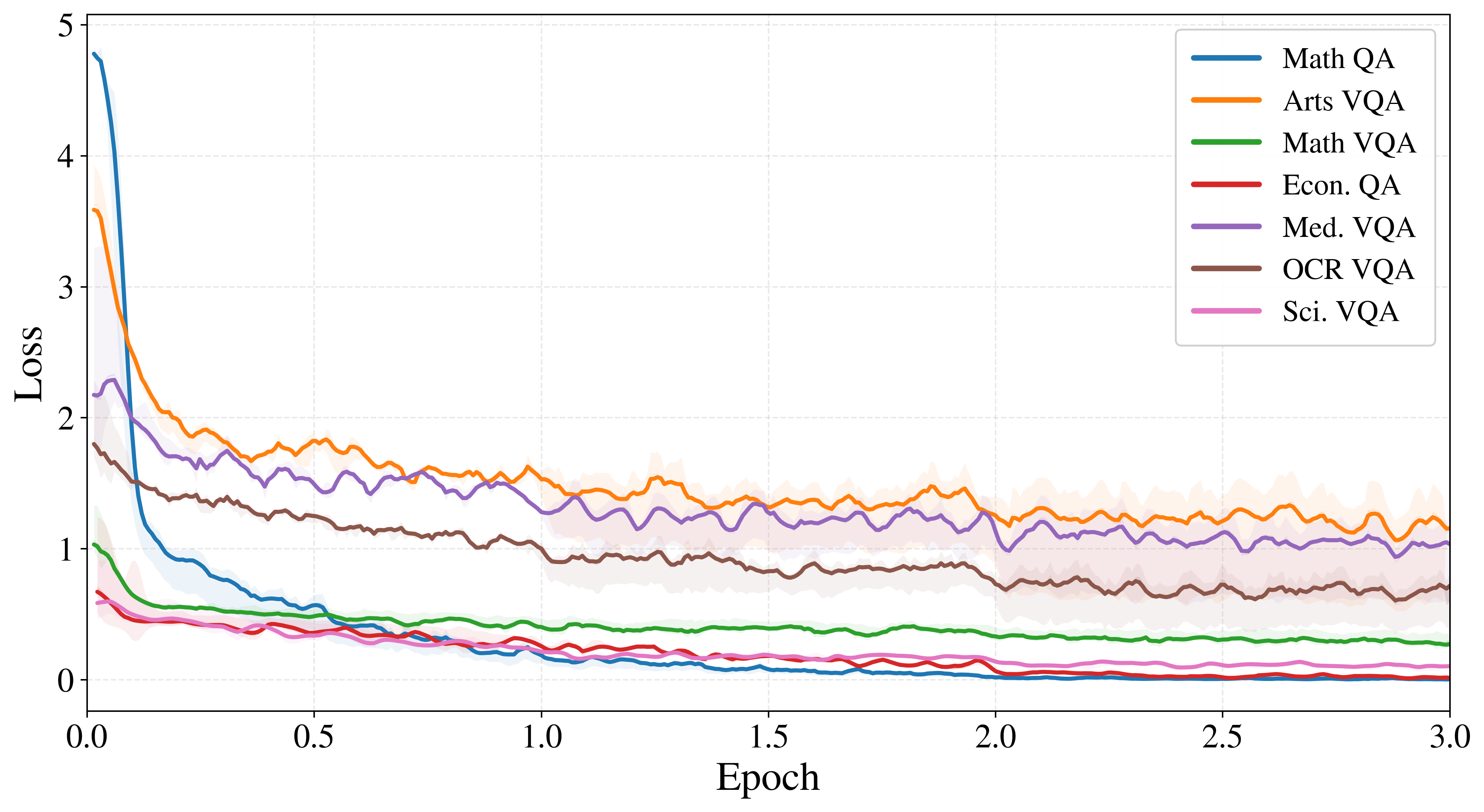}
  \caption{
  Training loss curves across tasks.
  }
  \label{fig:train_loss_compare}
\end{figure}

\subsection{Training Dynamics Across Tasks}
\label{sec:loss_dynamics}

To evaluate the robustness of our proposed PASs-MoE, we visualize its training loss curves on all downstream tasks from \textsc{MLLM-CTBench}. From Figure~\ref{fig:train_loss_compare}, we can observe clear task-dependent convergence rates, where some tasks start with higher loss and decrease more slowly than others. This disparity is consistent with the strong heterogeneity of \textsc{MLLM-CTBench} in terms of modality dependence and reasoning difficulty~\cite{guo2025mllm}.

Meanwhile, we can also notice that the optimization of our PASs-MoE on \textsc{MLLM-CTBench} remains relatively stable throughout the training procedure. That is, we do not observe any loss explosion phenomenon, and the loss of each task continues to decrease steadily during the later stages of training rather than rebounding. These discoveries suggest that our sequential training procedure does not suffer from severe optimization instability, and we do not observe evident overfitting under the adopted training protocol.
\subsection{In-depth Analysis}
\label{sec:analysis}
\paragraph{Routing drift correlates with old-task degradation.}
To examine whether the specialization between old-task inputs and experts is preserved under sequential updates, we track the gating distribution $G(x)$ for a fixed set of old-task inputs at two checkpoints: immediately after learning the corresponding task and after completing the full task stream. We quantify the change using distributional divergences (e.g., Jensen--Shannon divergence), averaged over layers and samples.
As shown in Fig.~\ref{fig:acc-vs-stability}, the MoE-LoRA baseline exhibits substantially larger routing drift, which coincides with lower final accuracy on old tasks, whereas our method yields consistently smaller drift while retaining higher final performance.
These results support the motivation in Section~\ref{sec:intro} that the \emph{misaligned co-drift} issue can undermine previously established input-expert specialization.
\begin{figure}[t]
  \centering
  \includegraphics[width=\linewidth]{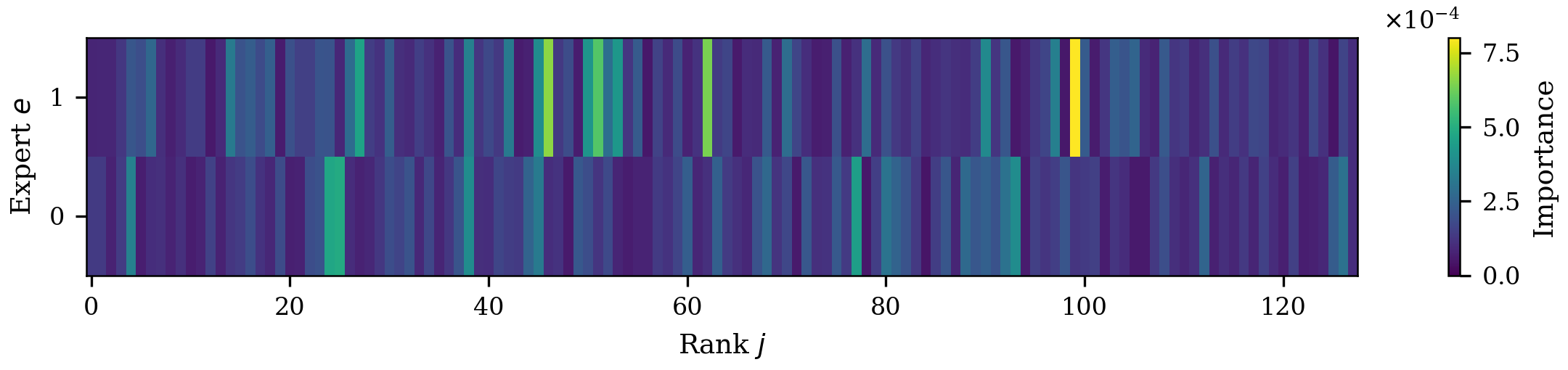}
  \caption{
\textbf{Aggregated rank importance over the task stream.}
We show the cumulative importance map $I_{\mathrm{agg}}$ over the sequential training procedure, and visualize the two most important experts for clarity. 
Importance concentrates on a small subset of all rank directions, suggesting that a few expert-rank directions dominate contributions across different tasks.
}
  \label{fig:Iagg-all}
\end{figure}

\begin{table}[t]
\centering
\begin{adjustbox}{max width=\columnwidth}
{\fontsize{9pt}{10pt}\selectfont
\setlength{\tabcolsep}{2.2mm}
\begin{tabular}{c c c c c c}
\toprule
\textbf{Method} &
\textbf{Experts} &
\textbf{Train Time} &
\textbf{Infer Time} &
\textbf{AP} &
\textbf{BWT} \\
\midrule

\rowcolor{gray!6}
\textbf{LoRA} & 1 & 120.14 & 129.87 &
28.24 & \forgetval{$-22.09$} \\
\midrule

\rowcolor{gray!6}
\textbf{MoE-LoRA} & 2 & 130.66 & 143.81 &
39.44 & \forgetval{$-10.18$} \\

\rowcolor{gray!10}
\textbf{Ours} & 2 & 140.39 & 132.78 &
46.61 & \forgetval{$-2.44$} \\
\midrule

\rowcolor{gray!6}
\textbf{MoE-LoRA} & 4 & 130.48 & 131.15 &
41.11 & \forgetval{$-7.91$} \\

\rowcolor{gray!10}
\textbf{Ours} & 4 & 140.29 & 148.59 &
44.85 & \forgetval{$-5.27$} \\
\midrule

\rowcolor{gray!6}
\textbf{MoE-LoRA} & 6 & 130.65 & 127.56 &
43.36 & \forgetval{$-6.64$} \\

\rowcolor{gray!10}
\textbf{Ours} & 6 & 140.89 & 132.84 &
\bestacc{48.46} & \secondacc{\forgetval{$-2.15$}} \\
\midrule

\rowcolor{gray!6}
\textbf{MoE-LoRA} & 8 & 130.60 & 130.91 &
43.58 & \forgetval{$-6.79$} \\

\rowcolor{gray!10}
\textbf{Ours} & 8 & 140.48 & 145.71 &
\secondacc{47.67} & \bestacc{\forgetval{$-2.13$}} \\
\bottomrule
\end{tabular}}
\end{adjustbox}

\caption{Effect of the number of experts on sequential multimodal continual learning. We compare our method against MoE-LoRA with softmax weighting across different expert counts. }
\label{tab:ablation_num_experts}
\end{table}

\paragraph{Activation-derived importance is highly non-uniform across ranks.}
Rank importance is highly non-uniform. PASs-RS uses activation statistics to localize which rank directions within each expert are most critical for retaining prior-task behavior. Intuitively, if a channel $(e,k)$ consistently yields a stronger low-rank response on earlier tasks, it is more likely to encode functionality that should be preserved and thus warrants stronger stabilization. To quantify this effect, we continuously record and accumulate an importance score $I^{\mathrm{agg}}_{e,k}$ based on the low-rank response $z_e = A_e h$. As shown in Fig.~\ref{fig:Iagg-all}, $I_{\mathrm{agg}}$ exhibits a sparse, concentrated pattern over the expert-rank grid rather than a uniform spread across rank directions, indicating that prior-task behavior is dominated by a small subset of rank directions. This pattern aligns with our PASs perspective that LoRA adaptation operates through structured, capability-tied pathways, making $I^{\mathrm{agg}}_{e,k}$ a principled rank-level weight for selectively stabilizing the most critical directions.

\paragraph{Rank stabilization reshapes updates toward low-importance directions.}
Building on the non-uniform importance, we examine whether rank-level stabilization preferentially protects high-importance directions.
For each layer, we sort rank directions by $I_{\mathrm{agg}}$, split them into quantile bins (e.g., deciles), and compute the median of $\log\|\Delta W\|$ with its interquartile range (IQR) within each bin.
As shown in Fig.~\ref{fig:upd_imp:repr} for representative layers (Layer~1/15/29), the median update magnitude generally decreases with increasing importance, indicating that stabilization suppresses drift on high-importance directions while leaving more flexibility to low-importance ones.
The trend strength varies across layers, consistent with layer-wise differences in representational roles and update dispersion: when importance is flatter or updates are more diffuse, the coupling between $I_{\mathrm{agg}}$ and $\Delta W$ becomes weaker.

\section{Conclusion}
\label{sec:conclusion}
We study CIT of MLLMs under a fixed-capacity Mixture-of-Experts adapter setting, where indiscriminate joint updates of routers and experts can trigger \emph{Misaligned Co-drift} and erode early input-expert specialization. We introduce \emph{PASs} induced by LoRA down-projections to provide a capability-tied coordinate system for aligning routing and preservation. Based on PASs, we propose a PASs-based MoE-LoRA method with PASs-guided Reweighting and PASs-aware Rank Stabilization, which couple routing to low-rank pathway responses and selectively stabilize historically important rank directions, respectively. On \textsc{MLLM-CTBench}, our method improves final performance and reduces forgetting, outperforming the second-best approach by \textbf{5.1\%} AP without increasing model capacity. We hope PASs offers a principled tool for capability-aligned routing and fine-grained preservation in CIT.

\section*{Limitations}

This work has several limitations. First, we focus on fixed-capacity MoE-LoRA in continual instruction tuning; the conclusions may not directly carry over to settings with capacity expansion, replay access, or substantially different continual learning paradigms. Second, PASs leverages the structural property of LoRA (the down-projection $A$ and the induced low-rank response $A_e h$), so extending the framework to other PEFT forms may require redefining a capability-aligned coordinate system and corresponding routing signals. Third, PASs-RW uses low-rank energy as a proxy for input-expert compatibility; while capability-related, its robustness and calibration under more severe distribution shifts or more complex instructions deserve further study. For PASs-RS, importance is estimated from activation statistics and can be noisy when task data are scarce or highly conflicting; it also requires maintaining historical statistics and tuning an additional regularization hyperparameters, which introduces extra implementation and tuning overhead. Finally, our evaluation primarily relies on AP/BWT, which captures overall performance and forgetting, but more direct measurements of routing mismatch, stronger interpretability analyzes of evolving expert responsibilities, and evaluations under realistic online distribution dynamics remain important future directions.

\section*{Acknowledgments}
This work was supported by the Beijing Natural Science Foundation (Grant No.~L252035).

\bibliography{custom}

@misc{HyTuning,
      title={Less Approximates More: Harmonizing Performance and Confidence Faithfulness via Hybrid Post-Training for High-Stakes Tasks}, 
      author={Haokai Ma and Lee Yan Zhen and Gang Yang and Yunshan Ma and Ee-Chien Chang and Tat-Seng Chua},
      year={2026},
      archivePrefix={arXiv},
      primaryClass={cs.LG},
      url={https://arxiv.org/abs/2604.08454}, 
}

@article{he2026plume,
  title={PLUME: Latent Reasoning Based Universal Multimodal Embedding},
  author={He, Chenwei and Hao, Xiangzhao and Yang, Tianyu and Ma, Yuxiang and Jia, Yuheng and Wu, Lingxiang and Zhao, Chaoyang and Guo, Haiyun and Wang, Jinqiao},
  journal={arXiv preprint arXiv:2604.02073},
  year={2026}
}

@inproceedings{BalDRO,
author = {Shao, Pengyang and Zhai, Naixin and Chen, Lei and Yang, Yonghui and Zhu, Fengbin and Yang, Xun and Wang, Meng},
title = {BalDRO: A Distributionally Robust Optimization based Framework for Large Language Model Unlearning},
year = {2026},
url = {https://doi.org/10.1145/3774904.3792975},
booktitle = {Proceedings of the ACM Web Conference 2026},
pages = {8874–8884},
series = {WWW '26}
}

@article{yang2026revisiting,
  title={Revisiting Robustness for LLM Safety Alignment via Selective Geometry Control},
  author={Yang, Yonghui and Tao, Wenjian and Liu, Jilong and Zhu, Xingyu and Fang, Junfeng and Huang, Weibiao and Wu, Le and Hong, Richang and Chua, Tat-Sent},
  journal={arXiv preprint arXiv:2602.07340},
  year={2026}
}

@article{NS4RS,
author = {Ma, Haokai and Xie, Ruobing and Meng, Lei and Feng, Fuli and Du, Xiaoyu and Sun, Xingwu and Kang, Zhanhui and Meng, Xiangxu},
title = {Negative Sampling in Recommendation: A Survey and Future Directions},
year = {2026},
url = {https://doi.org/10.1145/3793855},
doi = {10.1145/3793855},
journal = {ACM Trans. Inf. Syst.},
}

@article{zhang2025parameter,
  title={Parameter-efficient fine-tuning for foundation models},
  author={Zhang, Dan and Feng, Tao and Xue, Lilong and Wang, Yuandong and Dong, Yuxiao and Tang, Jie},
  journal={arXiv preprint arXiv:2501.13787},
  year={2025}
}

@article{sun2025stronger,
  title={A stronger mixture of low-rank experts for fine-tuning foundation models},
  author={Sun, Mengyang and Wang, Yihao and Feng, Tao and Zhang, Dan and Zhu, Yifan and Tang, Jie},
  journal={arXiv preprint arXiv:2502.15828},
  year={2025}
}

@article{Kirkpatrick_2017,
   title={Overcoming catastrophic forgetting in neural networks},
   volume={114},
   ISSN={1091-6490},
   url={http://dx.doi.org/10.1073/pnas.1611835114},
   DOI={10.1073/pnas.1611835114},
   number={13},
   journal={Proceedings of the National Academy of Sciences},
   publisher={Proceedings of the National Academy of Sciences},
   author={Kirkpatrick, James and Pascanu, Razvan and Rabinowitz, Neil and Veness, Joel and Desjardins, Guillaume and Rusu, Andrei A. and Milan, Kieran and Quan, John and Ramalho, Tiago and Grabska-Barwinska, Agnieszka and Hassabis, Demis and Clopath, Claudia and Kumaran, Dharshan and Hadsell, Raia},
   year={2017},
   month=mar, pages={3521–3526} }

@article{hu2022lora,
  title={Lora: Low-rank adaptation of large language models.},
  author={Hu, Edward J and Shen, Yelong and Wallis, Phillip and Allen-Zhu, Zeyuan and Li, Yuanzhi and Wang, Shean and Wang, Lu and Chen, Weizhu and others},
  journal={ICLR},
  volume={1},
  number={2},
  pages={3},
  year={2022}
}

@article{li2017learning,
  title={Learning without forgetting},
  author={Li, Zhizhong and Hoiem, Derek},
  journal={IEEE transactions on pattern analysis and machine intelligence},
  volume={40},
  number={12},
  pages={2935--2947},
  year={2017},
  publisher={IEEE}
}

@article{rolnick2019experience,
  title={Experience replay for continual learning},
  author={Rolnick, David and Ahuja, Arun and Schwarz, Jonathan and Lillicrap, Timothy and Wayne, Gregory},
  journal={Advances in neural information processing systems},
  volume={32},
  year={2019}
}

@inproceedings{farajtabar2020orthogonal,
  title={Orthogonal gradient descent for continual learning},
  author={Farajtabar, Mehrdad and Azizan, Navid and Mott, Alex and Li, Ang},
  booktitle={International conference on artificial intelligence and statistics},
  pages={3762--3773},
  year={2020},
  organization={PMLR}
}

@article{hao2026clear,  title={CLEAR: Unlocking Generative Potential for Degraded Image Understanding in Unified Multimodal Models},  author={Hao, Xiangzhao and Zhang, Zefeng and Zhang, Zhenyu and Yu, Linhao and Chen, Yao and Zhang, Yiqian and Guo, Haiyun and Wang, Shuohuan and Sun, Yu},  journal={arXiv preprint arXiv:2604.04780},  year={2026}}

@misc{llava,
      title={Improved Baselines with Visual Instruction Tuning}, 
      author={Haotian Liu and Chunyuan Li and Yuheng Li and Yong Jae Lee},
      year={2024},
      eprint={2310.03744},
      archivePrefix={arXiv},
      primaryClass={cs.CV},
      url={https://arxiv.org/abs/2310.03744}, 
}

@article{liu2023visual,
  title={Visual instruction tuning},
  author={Liu, Haotian and Li, Chunyuan and Wu, Qingyang and Lee, Yong Jae},
  journal={Advances in neural information processing systems},
  volume={36},
  pages={34892--34916},
  year={2023}
}

@inproceedings{radford2021learning,
  title={Learning transferable visual models from natural language supervision},
  author={Radford, Alec and Kim, Jong Wook and Hallacy, Chris and Ramesh, Aditya and Goh, Gabriel and Agarwal, Sandhini and Sastry, Girish and Askell, Amanda and Mishkin, Pamela and Clark, Jack and others},
  booktitle={International conference on machine learning},
  pages={8748--8763},
  year={2021},
  organization={PmLR}
}

@misc{Qwen2.5VL,
      title={Qwen2.5-VL Technical Report}, 
      author={Shuai Bai and Keqin Chen and Xuejing Liu and Jialin Wang and Wenbin Ge and Sibo Song and Kai Dang and Peng Wang and Shijie Wang and Jun Tang and Humen Zhong and Yuanzhi Zhu and Mingkun Yang and Zhaohai Li and Jianqiang Wan and Pengfei Wang and Wei Ding and Zheren Fu and Yiheng Xu and Jiabo Ye and Xi Zhang and Tianbao Xie and Zesen Cheng and Hang Zhang and Zhibo Yang and Haiyang Xu and Junyang Lin},
      year={2025},
      eprint={2502.13923},
      archivePrefix={arXiv},
      primaryClass={cs.CV},
      url={https://arxiv.org/abs/2502.13923}, 
}

@article{ewc,
  author       = {Abhishek Aich},
  title        = {Elastic Weight Consolidation {(EWC):} Nuts and Bolts},
  journal      = {CoRR},
  volume       = {abs/2105.04093},
  year         = {2021},
  url          = {https://arxiv.org/abs/2105.04093},
  eprinttype    = {arXiv},
  eprint       = {2105.04093},
  bibsource    = {dblp computer science bibliography, https://dblp.org}
}

@misc{lwf,
      title={Learning without Forgetting}, 
      author={Zhizhong Li and Derek Hoiem},
      year={2017},
      eprint={1606.09282},
      archivePrefix={arXiv},
      primaryClass={cs.CV},
      url={https://arxiv.org/abs/1606.09282}, 
}

@article{guo2025hide,
  title={Hide-llava: Hierarchical decoupling for continual instruction tuning of multimodal large language model},
  author={Guo, Haiyang and Zeng, Fanhu and Xiang, Ziwei and Zhu, Fei and Wang, Da-Han and Zhang, Xu-Yao and Liu, Cheng-Lin},
  journal={arXiv preprint arXiv:2503.12941},
  year={2025}
}

@misc{l2p,
      title={Learning to Prompt for Continual Learning}, 
      author={Zifeng Wang and Zizhao Zhang and Chen-Yu Lee and Han Zhang and Ruoxi Sun and Xiaoqi Ren and Guolong Su and Vincent Perot and Jennifer Dy and Tomas Pfister},
      year={2022},
      eprint={2112.08654},
      archivePrefix={arXiv},
      primaryClass={cs.LG},
      url={https://arxiv.org/abs/2112.08654}, 
}

@misc{replay,
      title={Experience Replay for Continual Learning}, 
      author={David Rolnick and Arun Ahuja and Jonathan Schwarz and Timothy P. Lillicrap and Greg Wayne},
      year={2019},
      eprint={1811.11682},
      archivePrefix={arXiv},
      primaryClass={cs.LG},
      url={https://arxiv.org/abs/1811.11682}, 
}

@misc{der,
      title={DER: Dynamically Expandable Representation for Class Incremental Learning}, 
      author={Shipeng Yan and Jiangwei Xie and Xuming He},
      year={2021},
      eprint={2103.16788},
      archivePrefix={arXiv},
      primaryClass={cs.CV},
      url={https://arxiv.org/abs/2103.16788}, 
}

@misc{mas,
      title={Memory Aware Synapses: Learning what (not) to forget}, 
      author={Rahaf Aljundi and Francesca Babiloni and Mohamed Elhoseiny and Marcus Rohrbach and Tinne Tuytelaars},
      year={2018},
      eprint={1711.09601},
      archivePrefix={arXiv},
      primaryClass={cs.CV},
      url={https://arxiv.org/abs/1711.09601}, 
}

@article{Chung2022ScalingIL,
  title={Scaling Instruction-Finetuned Language Models},
  author={Hyung Won Chung and Le Hou and S. Longpre and Barret Zoph and Yi Tay and William Fedus and Eric Li and Xuezhi Wang and Mostafa Dehghani and Siddhartha Brahma and Albert Webson and Shixiang Shane Gu and Zhuyun Dai and Mirac Suzgun and Xinyun Chen and Aakanksha Chowdhery and Dasha Valter and Sharan Narang and Gaurav Mishra and Adams Wei Yu and Vincent Zhao and Yanping Huang and Andrew M. Dai and Hongkun Yu and Slav Petrov and Ed Huai-hsin Chi and Jeff Dean and Jacob Devlin and Adam Roberts and Denny Zhou and Quoc V. Le and Jason Wei},
  journal={ArXiv},
  year={2022},
  volume={abs/2210.11416},
  url={https://api.semanticscholar.org/CorpusID:253018554 }
}

@inproceedings{razdaibiedina2023progressive,
  title={Progressive Prompts: Continual Learning for Language Models},
  author={Razdaibiedina, Anastasia and Mao, Yuning and Hou, Rui and Khabsa, Madian and Lewis, Mike and Almahairi, Amjad},
  booktitle={The Eleventh International Conference on Learning Representations},
  year={2023}
}

@article{guo2025mllm,
  title={MLLM-CBench: A Comprehensive Benchmark for Continual Instruction Tuning of Multimodal LLMs with Chain-of-Thought Reasoning Analysis},
  author={Guo, Haiyun and Hou, ZhiYan and Chen, Yu and He, Jinghan and Sun, Yandu and Zhou, Yuzhe and Guo, Shujing and Zhu, Kuan and Wang, Jinqiao},
  journal={arXiv preprint arXiv:2508.08275},
  year={2025}
}

@article{huang2024lg,
  title={Lg-cav: Train any concept activation vector with language guidance},
  author={Huang, Qihan and Song, Jie and Xue, Mengqi and Zhang, Haofei and Hu, Bingde and Wang, Huiqiong and Jiang, Hao and Wang, Xingen and Song, Mingli},
  journal={Advances in Neural Information Processing Systems},
  volume={37},
  pages={39522--39551},
  year={2024}
}

@article{magistri2024elastic,
  title={Elastic feature consolidation for cold start exemplar-free incremental learning},
  author={Magistri, Simone and Trinci, Tomaso and Soutif-Cormerais, Albin and van de Weijer, Joost and Bagdanov, Andrew D},
  journal={arXiv preprint arXiv:2402.03917},
  year={2024}
}

@article{saha2021gradient,
  title={Gradient projection memory for continual learning},
  author={Saha, Gobinda and Garg, Isha and Roy, Kaushik},
  journal={arXiv preprint arXiv:2103.09762},
  year={2021}
}

@article{roy2023subspace,
  title={Subspace distillation for continual learning},
  author={Roy, Kaushik and Simon, Christian and Moghadam, Peyman and Harandi, Mehrtash},
  journal={Neural Networks},
  volume={167},
  pages={65--79},
  year={2023},
  publisher={Elsevier}
}

@article{he2024seekr,
  title={Seekr: Selective attention-guided knowledge retention for continual learning of large language models},
  author={He, Jinghan and Guo, Haiyun and Zhu, Kuan and Zhao, Zihan and Tang, Ming and Wang, Jinqiao},
  journal={arXiv preprint arXiv:2411.06171},
  year={2024}
}

@article{hao2026trace,
  title={TRACE: Task-Adaptive Reasoning and Representation Learning for Universal Multimodal Retrieval},
  author={Hao, Xiangzhao and Wang, Shijie and Yang, Tianyu and Wang, Tianyue and Guo, Haiyun and Wang, Jinqiao},
  journal={arXiv preprint arXiv:2603.02929},
  year={2026}
}

@article{achiam2023gpt,
  title={Gpt-4 technical report},
  author={Achiam, Josh and Adler, Steven and Agarwal, Sandhini and Ahmad, Lama and Akkaya, Ilge and Aleman, Florencia Leoni and Almeida, Diogo and Altenschmidt, Janko and Altman, Sam and Anadkat, Shyamal and others},
  journal={arXiv preprint arXiv:2303.08774},
  year={2023}
}

@article{wenkmann2025variability,
  title={On The Variability of Concept Activation Vectors},
  author={Wenkmann, Julia and Garreau, Damien},
  journal={arXiv preprint arXiv:2509.24058},
  year={2025}
}

@article{zhang2025mm,
  title={MM-AttacKG: A Multimodal Approach to Attack Graph Construction with Large Language Models},
  author={Zhang, Yongheng and Zhao, Xinyun and Ma, Yunshan and Ma, Haokai and Guan, Yingxiao and Yang, Guozheng and Lu, Yuliang and Wang, Xiang},
  journal={arXiv preprint arXiv:2506.16968},
  year={2025}
}

@article{ma2025attackseqbench,
  title={AttackSeqBench: Benchmarking Large Language Models in Analyzing Attack Sequences within Cyber Threat Intelligence},
  author={Ma, Haokai and Yong, Javier and Ma, Yunshan and Kuei, Chen and Yusof, Anis and Liang, Zhenkai and Chang, Ee-Chien},
  year={2025}
}

@misc{CPRec,
  title={Large Language Model Empowered Recommendation Meets All-domain Continual Pre-Training}, 
  author={Haokai Ma and Yunshan Ma and Ruobing Xie and Lei Meng and Jialie Shen and Xingwu Sun and Zhanhui Kang and Tat-Seng Chua},
  year={2025},
  eprint={2504.08949},
  archivePrefix={arXiv},
  primaryClass={cs.IR},
  url={https://arxiv.org/abs/2504.08949}, 
}

@inproceedings{zhang2024improving,
  title={Improving Continual Few-shot Relation Extraction through Relational Knowledge Distillation and Prototype Augmentation},
  author={Zhang, Zhiheng and Zeng, Daojian and Bai, Xue},
  booktitle={Proceedings of the 2024 Joint International Conference on Computational Linguistics, Language Resources and Evaluation (LREC-COLING 2024)},
  pages={8756--8767},
  year={2024}
}

@inproceedings{kim2018interpretability,
  title={Interpretability beyond feature attribution: Quantitative testing with concept activation vectors (tcav)},
  author={Kim, Been and Wattenberg, Martin and Gilmer, Justin and Cai, Carrie and Wexler, James and Viegas, Fernanda and others},
  booktitle={International conference on machine learning},
  pages={2668--2677},
  year={2018},
  organization={PMLR}
}

@inproceedings{wang2023orthogonal,
  title={Orthogonal subspace learning for language model continual learning},
  author={Wang, Xiao and Chen, Tianze and Ge, Qiming and Xia, Han and Bao, Rong and Zheng, Rui and Zhang, Qi and Gui, Tao and Huang, Xuan-Jing},
  booktitle={Findings of the Association for Computational Linguistics: EMNLP 2023},
  pages={10658--10671},
  year={2023}
}

@article{jacobs1991adaptive,
  title={Adaptive mixtures of local experts},
  author={Jacobs, Robert A and Jordan, Michael I and Nowlan, Steven J and Hinton, Geoffrey E},
  journal={Neural computation},
  volume={3},
  number={1},
  pages={79--87},
  year={1991},
  publisher={MIT Press}
}

@article{shazeer2017outrageously,
  title={Outrageously large neural networks: The sparsely-gated mixture-of-experts layer},
  author={Shazeer, Noam and Mirhoseini, Azalia and Maziarz, Krzysztof and Davis, Andy and Le, Quoc and Hinton, Geoffrey and Dean, Jeff},
  journal={arXiv preprint arXiv:1701.06538},
  year={2017}
}

@misc{CoIN,
      title={CoIN: A Benchmark of Continual Instruction tuNing for Multimodel Large Language Model}, 
      author={Cheng Chen and Junchen Zhu and Xu Luo and Hengtao Shen and Lianli Gao and Jingkuan Song},
      year={2024},
      eprint={2403.08350},
      archivePrefix={arXiv},
      primaryClass={cs.CV},
      url={https://arxiv.org/abs/2403.08350}, 
}

@article{zhang2021tip,
  title={Tip-adapter: Training-free clip-adapter for better vision-language modeling},
  author={Zhang, Renrui and Fang, Rongyao and Zhang, Wei and Gao, Peng and Li, Kunchang and Dai, Jifeng and Qiao, Yu and Li, Hongsheng},
  journal={arXiv preprint arXiv:2111.03930},
  year={2021}
}

@inproceedings{sung2022vl,
  title={Vl-adapter: Parameter-efficient transfer learning for vision-and-language tasks},
  author={Sung, Yi-Lin and Cho, Jaemin and Bansal, Mohit},
  booktitle={Proceedings of the IEEE/CVF conference on computer vision and pattern recognition},
  pages={5227--5237},
  year={2022}
}

@article{ge2025dynamic,
  title={Dynamic Mixture of Curriculum LoRA Experts for Continual Multimodal Instruction Tuning},
  author={Ge, Chendi and Wang, Xin and Zhang, Zeyang and Chen, Hong and Fan, Jiapei and Huang, Longtao and Xue, Hui and Zhu, Wenwu},
  journal={arXiv preprint arXiv:2506.11672},
  year={2025}
}

@article{zhao2025each,
  title={Each rank could be an expert: Single-ranked mixture of experts lora for multi-task learning},
  author={Zhao, Ziyu and Zhou, Yixiao and Zhang, Zhi and Zhu, Didi and Shen, Tao and Li, Zexi and Yang, Jinluan and Wang, Xuwu and Su, Jing and Kuang, Kun and others},
  journal={arXiv preprint arXiv:2501.15103},
  year={2025}
}

@article{zhang2025enhancing,
  title={Enhancing Multimodal Continual Instruction Tuning with BranchLoRA},
  author={Zhang, Duzhen and Ren, Yong and Li, Zhong-Zhi and Yu, Yahan and Dong, Jiahua and Li, Chenxing and Ji, Zhilong and Bai, Jinfeng},
  journal={arXiv preprint arXiv:2506.02041},
  year={2025}
}

@article{fedus2022switch,
  title={Switch transformers: Scaling to trillion parameter models with simple and efficient sparsity},
  author={Fedus, William and Zoph, Barret and Shazeer, Noam},
  journal={Journal of Machine Learning Research},
  volume={23},
  number={120},
  pages={1--39},
  year={2022}
}

@inproceedings{huai2025cl,
  title={CL-MoE: Enhancing Multimodal Large Language Model with Dual Momentum Mixture-of-Experts for Continual Visual Question Answering},
  author={Huai, Tianyu and Zhou, Jie and Wu, Xingjiao and Chen, Qin and Bai, Qingchun and Zhou, Ze and He, Liang},
  booktitle={Proceedings of the Computer Vision and Pattern Recognition Conference},
  pages={19608--19617},
  year={2025}
}

@article{he2023continual,
  title={Continual instruction tuning for large multimodal models},
  author={He, Jinghan and Guo, Haiyun and Tang, Ming and Wang, Jinqiao},
  journal={arXiv preprint arXiv:2311.16206},
  year={2023}
}

@inproceedings{yu2024boosting,
  title={Boosting continual learning of vision-language models via mixture-of-experts adapters},
  author={Yu, Jiazuo and Zhuge, Yunzhi and Zhang, Lu and Hu, Ping and Wang, Dong and Lu, Huchuan and He, You},
  booktitle={Proceedings of the IEEE/CVF Conference on Computer Vision and Pattern Recognition},
  pages={23219--23230},
  year={2024}
}

@article{wang2024relational,
  title={Relational experience replay: Continual learning by adaptively tuning task-wise relationship},
  author={Wang, Quanziang and Wang, Renzhen and Li, Yuexiang and Wei, Dong and Wang, Hong and Ma, Kai and Zheng, Yefeng and Meng, Deyu},
  journal={IEEE Transactions on Multimedia},
  volume={26},
  pages={9683--9698},
  year={2024},
  publisher={IEEE}
}

@article{cui2025cmoa,
  title={CMoA: Contrastive Mixture of Adapters for Generalized Few-Shot Continual Learning},
  author={Cui, Yawen and Zhao, Jian and Yu, Zitong and Cai, Rizhao and Wang, Xun and Jin, Lei and Kot, Alex C and Liu, Li and Li, Xuelong},
  journal={IEEE Transactions on Multimedia},
  year={2025},
  publisher={IEEE}
}

@article{zhang2025cooper,
  title={COOPER: A Unified Model for Cooperative Perception and Reasoning in Spatial Intelligence},
  author={Zhang, Zefeng and Hao, Xiangzhao and Tang, Hengzhu and Zhang, Zhenyu and Sheng, Jiawei and Li, Xiaodong and Li, Zhenyang and Gao, Li and Shi, Daiting and Yin, Dawei and others},
  journal={arXiv preprint arXiv:2512.04563},
  year={2025}
}

@article{yang2026recall,
  title={ReCALL: Recalibrating Capability Degradation for MLLM-based Composed Image Retrieval},
  author={Yang, Tianyu and He, ChenWei and Hao, Xiangzhao and Wang, Tianyue and Guo, Jiarui and Guo, Haiyun and Qu, Leigang and Wang, Jinqiao and Chua, Tat-Seng},
  journal={arXiv preprint arXiv:2602.01639},
  year={2026}
}

\clearpage
\appendix


\section{Appendix}
\label{sec:appendix}

This appendix provides complementary details and additional results to support the main text. 
Appendix~\ref{app:setup} describes the experimental setup, including dataset details, task order construction, training hyperparameters, and evaluation protocols.
Appendix~\ref{app:order} reports results on MLLM-CTBench under an alternative task order, evaluating the robustness of our method to task order.
Appendix~\ref{app:extra} presents an additional validation of PASs-RS by contrasting our structured regularization with a random regularization baseline, isolating the effect of stabilizing task-important rank directions.

\subsection{Experimental Setup}
\label{app:setup}

We conduct all experiments on \textsc{MLLM-CTBench}, a recently proposed benchmark for continual instruction tuning of multimodal large language models. It is designed to evaluate continual adaptation under reasoning-intensive, non-saturated tasks with explicit domain shifts, making it particularly suitable for diagnosing catastrophic forgetting in modern MLLMs.

\textsc{MLLM-CTBench} consists of seven tasks spanning six diverse domains, including Math, Economics, Science, Medicine, OCR, and Arts. As summarized in Table~\ref{tab:task_composition}, the benchmark covers both text-only QA and vision-language VQA settings, and integrates data from 16 public datasets. The tasks jointly stress symbolic reasoning, visual grounding, OCR robustness, and domain-specific knowledge. To avoid task dominance in continual learning, the training data size of each task is carefully controlled to a comparable scale.

For each task, \textsc{MLLM-CTBench} defines a canonical instruction template and a task-specific final-answer evaluation metric, as shown in Table~\ref{tab:supp_prompts_metrics}. These templates are consistently applied across all methods, ensuring protocol-consistent comparison. Depending on the task characteristics, final answers are evaluated using Exact Match or ROUGE-L, while reasoning traces are used only for analysis and are not directly optimized unless specified.

Overall, the heterogeneity in modality, reasoning format, and domain semantics introduces substantial distribution shifts across tasks, posing significant challenges for continual instruction tuning. In this work, we strictly follow the official task order and training protocol of \textsc{MLLM-CTBench}, without revisiting data from previous tasks. Additional details on task ordering and evaluation settings are provided in Appendix~\ref{app:setup}.

\begin{table*}[t]
\centering
\small
\begin{tabular}{l l c c}
\hline
\textbf{Task} & \textbf{Data Source} & \textbf{Train} & \textbf{Test} \\
              &                     & \textbf{(Text / Image)} & \textbf{(Text / Image)} \\
\hline
Math QA       & TRACE                         & 10K / 0      & 0.5K / 0 \\
Economics QA  & TRACE                         & 5K / 0       & 0.5K / 0 \\
Science VQA   & AI2D, ScienceQA               & 9K / 4K      & 1K / 0.5K \\
Math VQA      & IconQA, GeoQA, CHARTX, MMMU   & 8.3K / 8.3K  & 0.9K / 0.9K \\
Medicine VQA  & VQA-RAD, VQA-Med-2021, PMC-VQA, PathVQA & 9K / 6.9K & 1K / 1K \\
OCR VQA       & ChartOCR, CROHME, ESTVQA      & 12K / 12.1K  & 1.4K / 1.4K \\
Arts VQA      & AQUA                          & 9K / 7K      & 1K / 0.9K \\
\hline
\end{tabular}
\caption{Statistics of the MLLM-CTBench datasets.}
\label{tab:task_composition}
\end{table*}

\begin{table*}[t]
\centering
\setlength{\tabcolsep}{2.6mm}
\renewcommand{\arraystretch}{1.05}
\resizebox{\textwidth}{!}{%
\begin{tabular}{l l l}
\toprule
\textbf{Task} & \textbf{Instruction Prompt (Canonical)} & \textbf{Final-Answer Metric} \\
\midrule
Math QA & Solve the following math problem and give your reasoning, then give the answer. & Exact Match \\
Economics QA & Give your reasoning about the monetary policy stance, then answer with the option’s letter directly. & Exact Match \\
Science VQA & Give the reasoning process, then answer with the option’s letter directly. & Exact Match \\
Math VQA & Analyze the problem and give the solution; then answer with the option’s letter. & Exact Match / ROUGE-L \\
Medicine VQA & Analyze and give the reasoning process, then answer using a single word or phrase. & ROUGE-L \\
OCR VQA & Give the reasoning process for text recognition, then answer using a single word or phrase. & ROUGE-L \\
Arts VQA & Analyze the artwork and give a reasoning process, then answer briefly. & ROUGE-L \\
\bottomrule
\end{tabular}%
}
\caption{Canonical instruction prompts and metrics across tasks.}
\label{tab:supp_prompts_metrics}
\end{table*}

\subsection{Results under an Alternative Task Order}
\label{app:order}
\paragraph{Task Orders.}
\textsc{MLLM-CTBench} provides two official continual instruction tuning protocols with different task orders.
Order-A follows:
Math QA $\rightarrow$ Arts VQA $\rightarrow$ Math VQA $\rightarrow$ Economics QA $\rightarrow$ Medicine VQA $\rightarrow$ OCR VQA $\rightarrow$ Science VQA.
Order-B is the reverse of Order-A.

\paragraph{Supplementary Results on Order-B.}
All main experiments in the paper follow Order-A.
To further examine robustness to task ordering, we additionally evaluate all methods under Order-B.
As shown in Table~\ref{tab:ohther_order_continual_results}, our method achieves the best overall performance and improves the final AP by \textbf{9.46\%} over the second-best method, indicating that the gains are not tied to a specific task sequence.

\begin{table*}[t]
\centering
\begin{adjustbox}{max width=\textwidth}
{\fontsize{9pt}{10pt}\selectfont
\setlength{\tabcolsep}{1mm}
\begin{tabular}{l
                cc cc cc cc cc cc cc
                cc}
\toprule
\multirow{2}{*}{\textbf{Method}} &
\multicolumn{2}{c}{\textbf{Sci.\ VQA}} &
\multicolumn{2}{c}{\textbf{OCR VQA}} &
\multicolumn{2}{c}{\textbf{Med.\ VQA}} &
\multicolumn{2}{c}{\textbf{Econ.\ QA}} &
\multicolumn{2}{c}{\textbf{Math VQA}} &
\multicolumn{2}{c}{\textbf{Arts VQA}} &
\multicolumn{2}{c}{\textbf{Math QA}} &
\multirow{2}{*}{\textbf{AP}} & \multirow{2}{*}{\textbf{BWT}} \\
\cmidrule(lr){2-3}\cmidrule(lr){4-5}\cmidrule(lr){6-7}\cmidrule(lr){8-9}
\cmidrule(lr){10-11}\cmidrule(lr){12-13}\cmidrule(lr){14-15}
& Acc & Forget & Acc & Forget & Acc & Forget & Acc & Forget
& Acc & Forget & Acc & Forget & Acc & Forget &  &  \\
\midrule

\textbf{SeqFT} &
\underline{56.84} & \forgetval{$-28.63$} &
11.47 & \forgetval{$-14.41$} &
20.01 & \forgetval{$-16.48$} &
35.65 & \forgetval{$-37.48$} &
36.57 & \forgetval{$-14.87$} &
9.08  & \forgetval{$-24.99$} &
53.94 & -- &
33.93 & \forgetval{$-17.56$} \\

\textbf{SeqLoRA} &
43.70 & \forgetval{$-42.63$} &
9.19  & \forgetval{$-15.74$} &
17.54 & \forgetval{$-19.18$} &
31.97 & \forgetval{$-40.51$} &
29.89 & \forgetval{$-23.93$} &
11.62 & \forgetval{$-25.76$} &
\textbf{55.17} & -- &
28.44 & \forgetval{$-23.96$} \\

\textbf{Replay} &
53.63 & \forgetval{$-28.93$} &
5.68  & \forgetval{$-16.74$} &
\underline{27.81} & \forgetval{$-5.87$} &
\underline{61.29} & \forgetval{$-11.49$} &
35.92 & \forgetval{$-11.40$} &
23.41 & \forgetval{$-10.50$} &
52.71 & -- &
\underline{37.21} & \forgetval{$-12.13$} \\

\textbf{L2P} &
44.33 & \forgetval{$-22.87$} &
9.47  & \forgetval{$-19.58$} &
19.68 & \forgetval{$-11.33$} &
39.54 & \forgetval{$-29.21$} &
31.36 & \forgetval{$-12.89$} &
6.87 & \forgetval{$-12.54$} &
51.45 & -- &
28.96 & \forgetval{$-15.49$} \\

\textbf{EWC} &
50.99 & \forgetval{$-31.20$} &
2.27  & \forgetval{$-20.17$} &
22.93 & \forgetval{$-12.09$} &
37.90 & \forgetval{$-35.29$} &
30.56 & \forgetval{$-22.12$} &
2.27  & \forgetval{$-31.25$} &
53.94 & -- &
28.69 & \forgetval{$-21.74$} \\

\textbf{LwF} &
52.40 & \forgetval{$-30.35$} &
\textbf{19.84} & \forgetval{$-1.95$} &
21.77 & \forgetval{$-11.25$} &
32.16 & \forgetval{$-39.01$} &
36.36 & \forgetval{$-15.07$} &
5.21  & \forgetval{$-28.42$} &
45.10 & -- &
30.41 & \forgetval{$-18.00$} \\

\textbf{MAS} &
51.08 & \forgetval{$-31.48$} &
2.84  & \forgetval{$-19.60$} &
22.70 & \forgetval{$-12.50$} &
37.10 & \forgetval{$-36.49$} &
30.22 & \forgetval{$-21.32$} &
2.63  & \forgetval{$-30.92$} &
53.20 & -- &
28.54 & \forgetval{$-21.76$} \\

\textbf{O-LoRA} &
\underline{56.84} & \forgetval{$-25.35$} &
13.49 & \forgetval{$-8.66$} &
23.63 & \forgetval{$-10.16$} &
34.53 & \forgetval{$-37.45$} &
35.48 & \forgetval{$-16.63$} &
\underline{29.16} & \forgetval{$-5.00$} &
\underline{54.84} & -- &
35.42 & \forgetval{$-14.75$} \\

\textbf{HiDe-LLaVA} &
46.65 & \forgetval{$-36.57$} &
\underline{15.17} & \forgetval{$-8.62$} &
24.87 & \forgetval{$-7.70$} &
47.78 & \forgetval{$-14.82$} &
\underline{43.33} & \forgetval{$-1.82$} &
28.33 & \forgetval{$-2.74$} &
42.12 & -- &
28.80 & \forgetval{$-16.99$} \\

\midrule

\rowcolor{gray!6}
\textbf{DDAS} &
50.05 & \forgetval{$-13.95$} &
1.07  & \forgetval{$-0.78$} &
21.65 & \forgetval{$-2.82$} &
54.64 & \forgetval{$-13.71$} &
37.06 & \forgetval{$+10.38$} &
13.48 & \forgetval{$-0.48$} &
40.89 & -- &
31.26 & \forgetval{$-3.05$} \\

\rowcolor{gray!6}
\textbf{MoELoRA (Top-k)} &
42.86 & \forgetval{$-35.62$} &
3.47  & \forgetval{$-18.32$} &
16.08 & \forgetval{$-16.39$} &
34.27 & \forgetval{$-32.92$} &
29.66 & \forgetval{$-19.08$} &
2.47  & \forgetval{$-30.61$} &
51.56 & -- &
25.77 & \forgetval{$-21.85$} \\

\rowcolor{gray!6}
\textbf{MoELoRA (Softmax)} &
50.42 & \forgetval{$-32.14$} &
2.27  & \forgetval{$-20.38$} &
22.94 & \forgetval{$-11.81$} &
38.71 & \forgetval{$-35.08$} &
31.24 & \forgetval{$-20.19$} &
2.49  & \forgetval{$-31.05$} &
53.45 & -- &
28.79 & \forgetval{$-21.52$} \\

\rowcolor{gray!10}
\textbf{Ours} &
\textbf{64.94} & \forgetval{$-17.72$} &
6.25  & \forgetval{$-16.04$} &
\textbf{28.11} & \forgetval{$-6.43$} &
\textbf{66.94} & \forgetval{$-1.61$} &
\textbf{46.52} & \forgetval{$-4.34$} &
\textbf{60.25} & \forgetval{$+26.99$} &
53.69 & -- &
\textbf{46.67} & \forgetval{$-2.74$} \\

\bottomrule
\end{tabular}}
\end{adjustbox}
\caption{Comparison with traditional methods and MoE-LoRA-based methods on \textsc{MLLM-CTBench}. Results are reported in the same order as the training sequence (from left to right). Best and second-best results for Acc and AP are marked in \textbf{bold} and \underline{underline}.}
\label{tab:ohther_order_continual_results}
\end{table*}

\begin{table*}[t]
\centering
\begin{adjustbox}{max width=\textwidth}
{\fontsize{9pt}{10pt}\selectfont
 \setlength{\tabcolsep}{1mm}
\renewcommand{\arraystretch}{1.08}
\begin{tabular}{c
                cc cc cc cc cc cc cc
                cc}
\toprule
\multirow{2}{*}{\textbf{Reg. scheme}} &
\multicolumn{2}{c}{\textbf{Math QA}} &
\multicolumn{2}{c}{\textbf{Arts VQA}} &
\multicolumn{2}{c}{\textbf{Math VQA}} &
\multicolumn{2}{c}{\textbf{Econ.\ QA}} &
\multicolumn{2}{c}{\textbf{Med.\ VQA}} &
\multicolumn{2}{c}{\textbf{OCR VQA}} &
\multicolumn{2}{c}{\textbf{Sci.\ VQA}} &
\multirow{2}{*}{\textbf{AP}} &
\multirow{2}{*}{\textbf{BWT}} \\
\cmidrule(lr){2-3}\cmidrule(lr){4-5}\cmidrule(lr){6-7}\cmidrule(lr){8-9}
\cmidrule(lr){10-11}\cmidrule(lr){12-13}\cmidrule(lr){14-15}
& Acc & Forget & Acc & Forget & Acc & Forget & Acc & Forget
& Acc & Forget & Acc & Forget & Acc & Forget &  &  \\
\midrule

\textbf{Random} &
49.01 & $-6.16$ &
40.79 & $+7.94$ &
41.50 & $-9.47$ &
57.10 & $-13.67$ &
27.50 & $-9.03$ &
20.19 & $-3.94$ &
79.76 & -- &
45.12 & $-4.90$ \\[0.6ex]

\textbf{Ours} &
49.52 & $-5.90$ &
43.22 & $+10.43$ &
44.70 & $-6.50$ &
66.13 & $-6.05$ &
29.99 & $-5.38$ &
21.95 & $-1.63$ &
83.73 & -- &
48.46 & $-2.15$ \\

\bottomrule
\end{tabular}}
\end{adjustbox}

\caption{Ablation on the regularization scheme in sequential multimodal continual learning. We compare random regularization with our projection-aware regularization while keeping all other components fixed. }
\label{tab:ablation_reg_scheme}
\end{table*}


\subsection{Comparison with Random Rank Regularization}
\label{app:extra}

To further validate the effectiveness of \emph{PASs-RS}, we compare it with a random rank regularization scheme under the same MoE-LoRA parameter budget. Specifically, instead of weighting the paired rank directions in $A$ and $B$ according to their historical importance in the pathway activation subspace, the random baseline assigns stabilization weights uniformly at random.

The results are reported in Table~\ref{tab:ablation_reg_scheme}. Random rank regularization consistently leads to inferior performance, indicating that constraining arbitrary rank directions does not effectively preserve task-relevant knowledge and may even hinder adaptation to new tasks. In contrast, PASs-RS yields more reliable improvements by selectively stabilizing rank directions that are important for previous tasks, as identified by activation statistics in the pathway activation subspace. This comparison highlights the importance of using capability-aligned signals, rather than indiscriminate or random constraints, when designing rank-level stabilization mechanisms for continual instruction tuning.

\end{document}